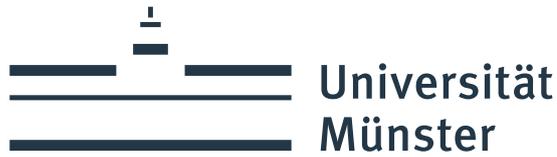

# Improving Automated Algorithm Selection by Advancing Fitness Landscape Analysis

## Inauguraldissertation

zur Erlangung des akademischen Grades

eines Doktors der Wirtschaftswissenschaften

durch die Wirtschaftswissenschaftliche Fakultät der

Universität Münster

vorgelegt von

## Raphael Patrick Prager

Münster, 12. Oktober 2023

# ACKNOWLEDGEMENTS

I would like to express my earnest gratitude to my supervisor Heike Trautmann for her guidance, inspiration, never-ending patience, and willingness to give feedback at ungodly hours. I am more than thankful to have been under her tutelage. My academic growth has been shaped by her guidance. I would not be here if it wasn't for her.

Furthermore, I would like to thank my colleagues Moritz Seiler, Janina Lütke Stockdiek, Laura Troost, Julia Seither, Pascal Kerschke, Jakob Bossek, Jeroen Rook, Lennart Schäpermeier, Christian Grimme, Lucas Stampe, Lena Clever, Ingolf Terveer, and Oliver Preuß for all the joyful times and joint work over the years. Whether it being inside or outside the office.

I am also thankful for my parents, who were there for me my entire life. Special thanks are due to my partner Mareike. Thank you for always having my back and supporting me through these 'interesting' times. Even at the expense of your own free time and nerves.

No thanks is directed to my best friends Henrik, Jonas, and Markus (ordered alphabetically of course). In fact, they should thank me for gracing them with my boundless musical talents, the joy and optimism I brought to their bleak lives, and the sheer effort it took me to keep this band of misfits together. All jokes aside, thank you guys for being through thick and thin over a decade and providing ample support (or distractions) in times of need. I had to keep it short; otherwise, certain people would not be happy with the ratio of allocated space.

Raphael Patrick Prager

# Abstract

Optimization is ubiquitous in our daily lives. In the past, (sub-)optimal solutions to any problem have been derived by trial and error, sheer luck, or the expertise of knowledgeable individuals. In our contemporary age, there thankfully exists a plethora of different algorithms that can find solutions more reliably than ever before. Yet, choosing an appropriate algorithm for any given problem is challenging in itself. The field of automated algorithm selection provides various approaches to tackle this latest problem. This is done by delegating the selection of a suitable algorithm for a given problem to a complex computer model. This computer model is generated through the use of Artificial Intelligence. Many of these computer models rely on some sort of information about the problem to make a reasonable selection. Various methods exist to provide this informative input to the computer model in the form of numerical data.

In this cumulative dissertation, I propose several improvements to the different variants of informative inputs. This in turn enhances and refines the current state-of-the-art of automated algorithm selection. Specifically, I identify and address current issues with the existing body of work to strengthen the foundation that future work builds upon. Furthermore, the rise of deep learning offers ample opportunities for automated algorithm selection. In several joint works, my colleagues and I developed and evaluated several different methods that replace the existing methods to extract an informative input. Lastly, automated algorithm selection approaches have been restricted to certain types of problems. I propose a method to extend the generation of informative inputs to other problem types and provide an outlook on further promising research directions.

# Contents

# Chapter 1

# Introduction

> *'Some problems are just too complicated for rational logical solutions. They admit of insights, not answers.'*
>
> Jerome Bert Wiesner

One of mankind's propensities is the pursuit of optimizing everything that holds even a minimal degree of significance. Whether it pertains to our individual lives where we, for example, try to find the fastest way to work, or to problems rooted in the corporate sector. The problems of the latter also encompass a diverse array of application areas. For instance, these optimization problems can manifest as creating optimal schedules for classes at a university, the development of cost-effective routes for delivery drivers, or the identification of a suitable set of parameters for a machine in a production line. Inherent to any of these problems is that they provide us a form of agency where we can exert influence over the eventual outcome through conscious decision-making. Many of these problems lack a known closed-form mathematical representation and thereby do not offer analytical gradient information. This effectively renders them, for all intent and purposes, a black-box. Thus, these problems are commonly referred to as *black-box optimization* problems.

In case of the machine situated in a production line, an individual might deliberate about the adjustment of the parameters of the machine. In the past, a good setting was (and to this day may even still be) created by trial-and-error and/or stemmed from the domain knowledge and expertise of proficient workers. While not axiomatically destined for failure, the efficacy of such an 'optimization procedure' is severely constrained since it either consumes a significant amount of resources or is reliant on a small group of individuals.

With the proliferation of technological advancements, we fortunately can delegate this task to sophisticated algorithms. Over the past decades, a myriad of algorithms have been developed and were made accessible to the general public. Yet, it can be asserted that no single algorithm is superior to all others. In other words, any given problem instance may necessitate a specific algorithm to be solved optimally in terms of the quality of the found solution, the expenditure of time, or resources to solve the problem instance. This phenomenon is often described as the performance complementary behavior between algorithms [21].

Ultimately, the issue – which any person trying to solve a particular black-box optimization problem encounters – has shifted from finding a viable solution to an optimization problem to identifying an appropriate algorithm to solve said optimization problem. This challenge has been formalized by [60] and is referred to as the *algorithm selection problem*. Providing a remedy to this problem can lead to a tangible reduction in time spent, energy consumption, and an increase in quality and quantity of found solutions.

Hence, this thesis pursues the generation of adequate computer-aided models that enable researchers and practitioners alike to find an appropriate algorithm automatically. Granted, there already exists a plethora of different approaches. Yet, the contributed material of this thesis addresses pressing issues with current solutions, explores new technologies, and generally widens the scope of existing solutions.

## 1.1 Algorithm Selection Problem

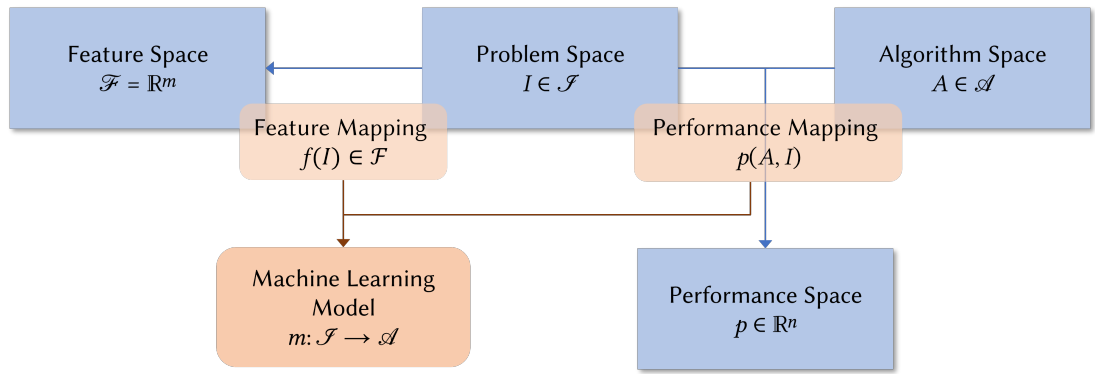

Figure 1.1: Extended illustration of the algorithm selection problem as proposed by [60]. It is supplemented by an additional node representing machine learning models for automated algorithm selection.

At its core, the algorithm selection problem (ASP) strives to determine a suitable algorithm $A \in \mathcal{A}$ out of a portfolio of algorithms $\mathcal{A}$ which constitutes the **algorithm space**. This selection happens on an instance level. Meaning, rather than finding a single algorithm for a set of problem instances of the **problem space** $\mathcal{I}$, the objective is to construct a mapping $m : \mathcal{I} \to \mathcal{A}$ which chooses an optimal algorithm $A$ for any given problem instance $I \in \mathcal{I}$. Generally, the problem space $\mathcal{I}$ covers a plethora of instances with varying degree of problem hardness and other criteria.

This optimality refers to some performance metric $p \in \mathbb{R}^n$. In the majority of cases, this metric is represented by a singular scalar but may contain multiple values. This metric is the result of the performance mapping $p : \mathscr{A} \times \mathscr{I} \to \mathbb{R}^n$ where a single algorithm $A$ is run on a specific problem instance $I$. The type of metric is dependent on the general type of problem. This can range from simple metrics such as the number of function evaluations or computational time to even the convergence speed throughout the search required to reach a satisfactory solution.

Indispensable for solutions to the ASP is the presence of some form of quantitative information about a problem instance. This general information is part of the **feature space** $\mathscr{F}$ and is derived independently from any algorithm. Essentially, it requires the existence of some procedure to extract this information of a given problem instance and that is called a **feature mapping** $f(I)$. In the scope of the ASP, this process produces a set of numerical features.

A prominent solution to the ASP is to automate the selection process for a given problem instance $I$. In academia, this is often referred to as *per-instance automated algorithm selection* (AAS) [21]. Several strategies have emerged over the past decades to tackle the ASP. Some solutions let multiple solvers run in parallel and terminate specific algorithms which exhibit a poor performance early to focus on more promising solvers (e.g. [33]). Admittedly, they discard the entire feature space so it is controversial to categorize them as AAS solutions. Other strategies - which are of more interest for this work - incorporate machine learning (ML) models to make informed decisions utilizing $f(I)$ [25, 21]. Based on a body of historic performance data of algorithms created through the performance mapping, we can train an ML model to learn the mapping $m$. The different introduced concepts, relationships and dependencies are depicted in Figure 1.1.

The quality of an AAS model is typically assessed by comparing it to the *single best solver* (SBS) and the *virtual best solver* (VBS). The SBS is part of the algorithm space $\mathscr{A}$. As the name already hints at, the SBS refers to a single algorithm $A \in \mathscr{A}$ which exhibits overall the best performance $\bar{p}$. Here, $\bar{p}$ represents an aggregated value which is calculated over a set of individual performance metrics via a suitable aggregation method (e.g., arithmetic mean). The VBS on the other hand is a hypothetical AAS model which always chooses the optimal algorithm $A \in \mathscr{A}$ for any given problem instance $I \in \mathscr{I}$. Essentially, it serves as a lower bound (for minimization problems) which every AAS model founded in reality strives to imitate. Closing the gap between the SBS and VBS via an AAS model is the objective.

One of the merits innate to AAS is its independence from the actual optimization domain. The previously introduced concepts can be applied to combinatorial problems (e.g. Traveling Salesman Problem) or to continuous, binary, integer, or a mixture thereof problems. In essence, any conceivable problem domain is viable. AAS is also agnostic to the number of objectives. Meaning, it can be applied to single-objective or multi-objective problems. This thesis primarily focuses on garnering further insights for AAS in the single-objective continuous black-box optimization domain with box-constraints. This is later extended to cover problems with mixed search spaces. The former can be defined as follows:

$$\begin{aligned} \min \quad & f(\mathbf{x}) \\ \text{s.t.} \quad & x_i^l \leq x_i \leq x_i^u \quad i = 1, \ldots, n, \end{aligned}$$

where $f : \mathbb{R}^n \to \mathbb{R}$ is the objective function to be minimized with a dimensionality of $n$. The vector $\mathbf{x} = (x_1, \ldots, x_n)^\top \in \mathbb{R}^n$ is an $n$-dimensional vector of decision variables of the decision space $\mathbb{R}^n$. The respective lower and upper bounds of the box-constraints for a decision variable $x_i$ are represented by $x_i^l$ and $x_i^u$.

For mixed-variable problems, the decision space changes to account for additional types of decision variables. It can be defined as $f : \mathbb{R}^n \times \mathbb{Z}^k \times \mathbb{Z}^p \to \mathbb{R}$. Here, $\mathbb{Z}^k$ is the $k$-dimensional space of $k$ integer decision variables. On the other hand, $\mathbb{Z}^p$ provides the space for $p$ categorical decision variables, which are also typically represented by integer values [46]. In this thesis, all considered problems are not subject to any other constraints than the box-constraints. Throughout this thesis, the total dimensionality of a problem is denoted as $D$.

The entire field of AAS has flourished in recent years where advances occurred in the algorithm space $\mathcal{A}$ through the proposal of new algorithms [20, 17], different interpretations of the mapping $m$ [6, 25, 51], as well as the development of more sophisticated procedures related to the feature space [23, 34, 43]. As this thesis is deeply rooted in advancing the latter for the *single-objective black-box optimization domain*, the following section is dedicated to elaborating the relevant elements of the feature space in more detail. This is further contextualized around existing limitations and open challenges.

## 1.2 Numerical Instance Features and Open Challenges

Dependent on the problem domain, the process of mapping a problem instance to a set of numerical features involves different techniques and is known under different names. In essence, the vast majority of methods aim to characterize the fitness landscape of a problem instance based on certain characteristics and is therefore often called fitness landscape analysis [36]. In the continuous domain, a different term has been widely accepted which originates from the works of [39, 38] and is called *Exploratory Landscape Analysis* (ELA).

In their preceding work which lead up to the development of numerical ELA features, the authors of [39] formalized eight of these landscape characteristics and called them high-level properties. These are the degree of multi-modality, variable scaling, separability, existence of plateaus, and global structure as well as the contrast between local and global optima, search space homogeneity, and basin size homogeneity. Under the premise that an algorithm exhibits a similar performance on two problem instances when these two are also very similar in terms of their landscape high-level properties, the authors of [38] developed a set of low-level features which can measure certain high-level properties of any given fitness landscape. In their work, they have demonstrated the capability of these low-level features by assigning the high-level properties of well-studied benchmark problems with a high success rate. The merits of numerical landscapes features has been further corroborated with the development of an AAS model making use of ELA features [6].

Further advances in that area since the inception of ELA have been made. This includes feature sets which measure the ruggedness of a fitness landscape [43] as well as features which detect the presence of funnels - a special type of global structure - in a fitness landscape [23]. However, this is only a small excerpt of various methods proposed throughout the last two decades. Several more exist such as the features sets proposed by [1, 3, 22, 34, 69].

Nevertheless, these features incur a cost which has to be accounted for in the performance evaluation of AAS models. Meaning, the basis for any feature calculation is a small well-distributed sample of the decision space and the respective objective values. This is often referred to as the *initial design*. Based on this initial design, several groups of low-level features can be computed. Yet, some feature sets require additional function evaluations which makes them less suited for AAS as this could have been used by an algorithm as additional budget. Therefore, any landscape features which are suitable for AAS have three inherent properties.

1. They are relatively cheap by only requiring a small initial design.
2. They possess discriminating power to distinguish between problem instances.
3. Their computation is fast.

Despite the irrefutable effectiveness of ELA in AAS [6, 25, 21], it is important to also evaluate them critically. Meaning, while the individual computation of features for a single problem instance might be considered fast, at scale it still requires substantial resources. In addition, ELA features are a hand-crafted product which required significant effort, insight, and development time. However, the research community as a whole still lacks insight and comprehension in some aspects of fitness landscape analysis. Hence, we cannot ensure that all the relevant high-level properties are captured within these features.

## 1.3 Contribution of the Thesis

Numerous works have already provided different solutions to the ASP via automated feature-based algorithm selection. Yet, there still remains a significant amount of unanswered questions that warrants further exploration of the field. During the course of my PhD, my research was driven by three research questions:

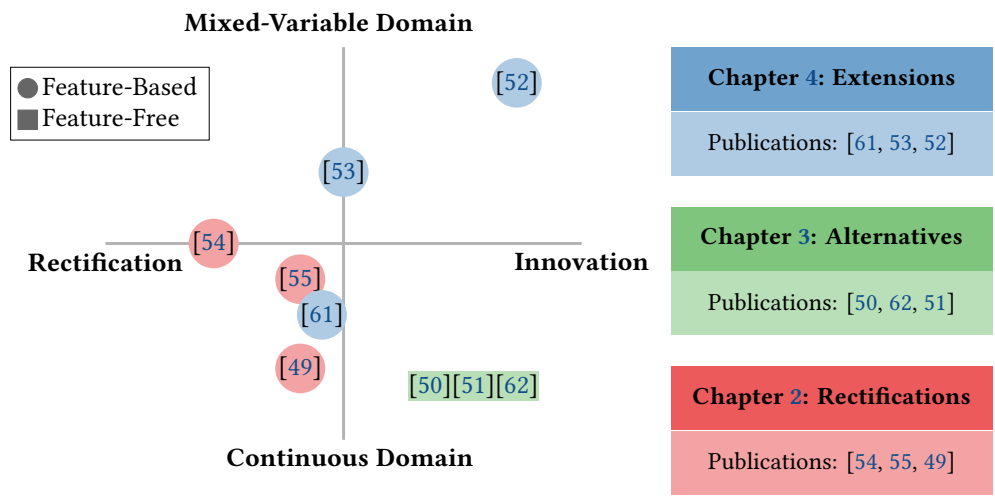

Figure 1.2: The left hand-side provides a classification of publications based on two continuums and a binary option. The right hand-side provides the structure of this thesis which is derived from this classification.

**RQ1.** What are shortcomings of current practices in the AAS and landscape research communities, and what are possible remedies?

**RQ2.** Is it possible to substitute or enhance existing feature-based approaches with novel technologies from adjacent research fields?

**RQ3.** Is it possible to disseminate the usage of ELA to other single-objective problem domains?

Throughout my academic career, I have published several works pertaining to different subjects within the field of AAS. However, all are related to at least one of the aforementioned research questions. These range from addressing current issues, gaps, or even misconceptions in the research community to providing alternative solutions to common practices and even expanding the scope of the state-of-the-art.

Nine of these publications are included in this thesis and can be classified according to the outlined criteria. The first spectrum used to categorize publications is based on whether they identify and address contemporary issues with the status quo or are more of an inventive nature. The second criterion can also be conceptualized on a continuum, with one end representing the purely continuous problem domain and the other end the mixed-variable problem space where the decision space can attain an arbitrary mixture of continuous, integer, binary, and categorical decision variables. The last dimension I am going to categorize my work on is whether it pertains to feature-based or feature-free approaches. These publications, in order as they are discussed in this thesis, are:

- [54] : Nullifying the Inherent Bias of Non-Invariant Exploratory Landscape Analysis Features (EvoApplications 2023)

- [55] : Pflacco: Feature-Based Landscape Analysis of Continuous and Constrained Optimization Problems in Python (ECJ 2023)

- [49] : Neural Networks as Black-Box Benchmark Functions Optimized for Exploratory Landscape Features (FOGA 2023)

- [50] : Towards Feature-Free Automated Algorithm Selection for Single-Objective Continuous Black-Box Optimization (SSCI 2021)

- [62] : A Collection of Deep Learning-based Feature-Free Approaches for Characterizing Single-Objective Continuous Fitness Landscapes (GECCO 2022)

- [51] : Automated Algorithm Selection in Single-Objective Continuous Optimization: A Comparative Study of Deep Learning and Landscape Analysis Methods (PPSN 2022)

- [61] : HPO×ELA: Investigating Hyperparameter Optimization Landscapes by Means of Exploratory Landscape Analysis (PPSN 2022)

- [53] : Investigating the Viability of Existing Exploratory Landscape Analysis Features for Mixed-Integer Problems (GECCO 2023)

- [52] : Exploratory Landscape Analysis for Mixed-Variable Problems (TEVC under Review)

A thematic arrangement according to the aforementioned dimensions is depicted in Figure 1.2. It should be noted that the veracity of the placement of each publication is a matter of subjective interpretation. From this, I derived the structure of this thesis, which organizes the publications into three chapters. In each chapter, the pertinent research questions are revisited and linked to the corresponding contents of that chapter. I commence with Chapter 2 in which I identify several issues with the current state of research in AAS along with possible remedies. Chapter 3, on the other hand, presents novel AAS solutions that still require an initial design but do not rely on any landscape features at all. In Chapter 4, several works are introduced which continuously work towards an extension of the problem domain to which ELA has usually been applied. Lastly, I conclude my thesis in Chapter 5, wherein I summarize the findings with respect to my research questions and provide an outlook on possible future research avenues.

Chapter 2

# Rectifying Current Issues in Automated Algorithm Selection

## Contents

## Addressed Research Question

**RQ1.** What are shortcomings of current practices in the AAS and landscape research communities, and what are possible remedies?

## Contributed Material

[54] Raphael Patrick Prager and Heike Trautmann. 'Nullifying the Inherent Bias of Non-Invariant Exploratory Landscape Analysis Features'. In: *Applications of Evolutionary Computation*. Ed. by João Correia, Stephen Smith and Raneem Qaddoura. Cham: Springer International Publishing, 2023, pp. 411–425. ISBN: 978-3-031-30229-9. DOI: 10.1007/978-3-031-30229-9_27.

[55] Raphael Patrick Prager and Heike Trautmann. 'Pflacco: Feature-Based Landscape Analysis of Continuous and Constrained Optimization Problems in Python'. In: *Evolutionary Computation* (2023), pp. 1–25. ISSN: 1063-6560. DOI: 10.1162/evco_a_00341.

[49]   Raphael Patrick Prager, Konstantin Dietrich, Lennart Schneider, Lennart Schäpermeier, Bernd Bischl, Pascal Kerschke, Heike Trautmann and Olaf Mersmann. 'Neural Networks as Black-Box Benchmark Functions Optimized for Exploratory Landscape Features'. In: *Proceedings of the 17th ACM/SIGEVO Conference on Foundations of Genetic Algorithms*. FOGA '23. Potsdam, Germany: Association for Computing Machinery, 2023, pp. 129–139. ISBN: 979-8-400-70202-0. DOI: 10.1145/3594805.3607136.

## 2.1   Nullifying the Inherent Bias of Non-Invariant Exploratory Landscape Analysis Features

Scholarly works, like for instance [34], [25], and [21], highlight the rapid development and utilization of ELA features within the optimization community. These works encompass the comprehension of different fitness landscapes, their peculiarities and impact on algorithm performance through the analysis of ELA features [58], but also evaluate which ELA features are most informative, i.e., most influential in distinguishing between fitness landscapes and thereby problem instances [59]. In [54], we identified a set of ELA features that are not invariant to linear transformations (i.e., shifts and scaling) of the objective space. This implies that this set of ELA features is primarily sensitive to the absolute values of any objective function and only secondarily dependent on the actual properties of the fitness landscape. This is an issue in our research field which we deem necessary to resolve.

This issue is particularly pronounced when examining the diverse black-box optimization benchmark (BBOB) [16]. BBOB comprises a set of 24 distinct optimization problems (called functions) in various dimensions. Each BBOB function can be subjected to certain variations, i.e., shifts, scaling, and rotations. Thereby, we produce several different instances of any given BBOB function while maintaining the general landscape characteristics. Based on these properties, BBOB is structured into five groups with similar landscape characteristics. Individual instances of BBOB functions also offer various and distinct ranges of objective values. These different ranges can be assessed in Figure 2.1 where we can construct a compartmentalization of problem instances (BBOB functions) solely based on the ranges of possible objective values.

Regrettably, the set of most informative ELA features (according to [59]) and the non-invariant feature set we discovered have clear overlaps. When these features are used for studies on BBOB or other diverse benchmark suites, the results are inherently biased. The non-invariant ELA features exploit and integrate information (about the range of objective values), which does not generalize beyond the given study. This is because the findings or models only adopt a limited understanding of the actual structural landscape properties of the respective problem instances. Exemplary non-invariant ELA features are `ela_meta.lin_simple.intercept` and `ic.eps.s`.

## 2.1 Nullifying the Inherent Bias of Non-Invariant Exploratory Landscape Analysis Features

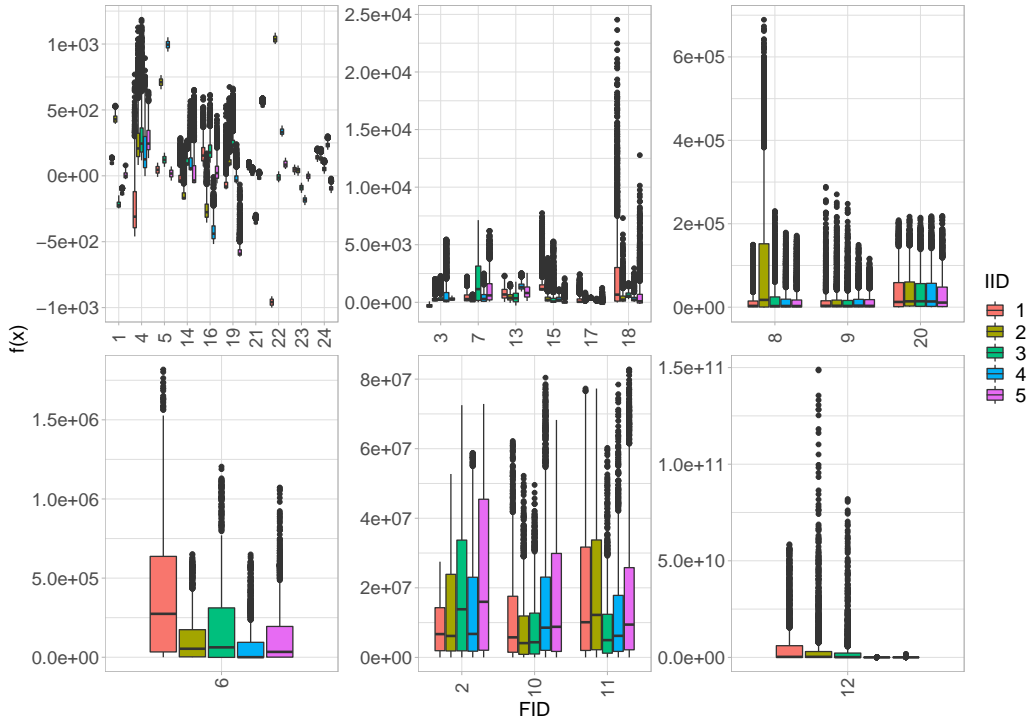

Figure 2.1: Six groups of BBOB functions and their first five respective instances. The groups are derived solely based on the range of objective values each problem instance exhibits. Those with similar ranges are grouped together. Figure taken from [54].

As a remedy, we proposed the usage of a straightforward yet elegant normalization technique to circumvent these effects. This method is to be applied on the objective values of the initial design prior to the computation of ELA features. This does not impose any restrictions or requirements beyond the information available in the initial design. In other words, we suggest to use 'min-max normalization' on the objective values to scale the values to [0, 1], irrespective of the actual scale of the original problem. While we prove that this procedure makes non-invariant ELA features invariant to scaling and shifting of problem instances, we also provide a juxtaposition of the previous computation of ELA features versus our recommended approach. This underscores the detrimental effects that prior works have been subjected to.

We place particular emphasis on the repercussions for AAS models. Here, we demonstrate the artificial superiority of models trained on previous ELA feature computations by leaving only some problem variations (BBOB IIDs) out. The significance of these advantages diminishes when we exclude entire problems (BBOB FIDs) from the training procedure. In the last experiment, we highlight the merits of our method compared to the traditional ELA feature computation by leaving out collections of problems based on the groupings of Figure 2.1. Our model is capable to generalize better to unseen objective value ranges simply because these are normalized to $[0, 1]$ whereas the other model loses its artificial advantage it had in previous settings.

## 2.2 Pflacco: Feature-Based Landscape Analysis of Continuous and Constrained Optimization Problems in Python

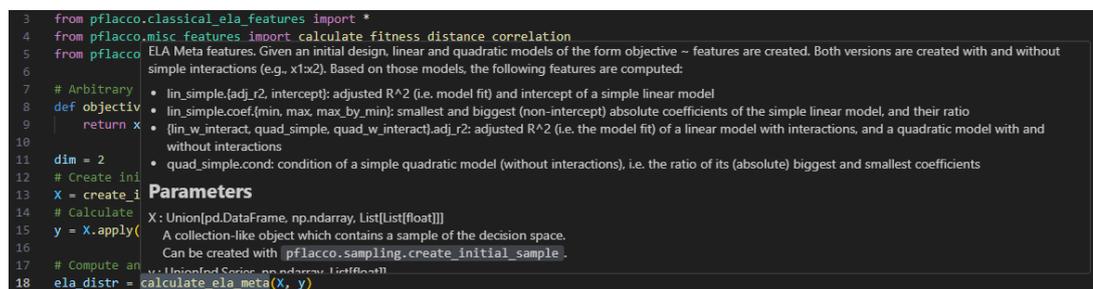

Figure 2.2: Screenshot of the inline documentation of `pflacco` inside of the integrated development environment 'Visual Studio Code'. This inline documentation can be perused in its entirety by scrolling down.

Making use of the plethora of proposed landscape features as well as keeping up with general advancements (e.g., the findings of the previous section) is a burden for researchers and practitioners. This is alleviated to some degree by the R-package `flacco` [26]. This package provides a comprehensive implementation of ELA features. However, the programming language R wavers in popularity compared to other emerging languages such as Python. This is especially true for newer generations of aspiring researchers. To make landscape analysis more accessible to a wider audience and reduce the barrier of entry, I implemented the majority of existing ELA features of `flacco` in a Python package called `pflacco` [55]. This work provides two additional contributions compared to the R version. Firstly, we systematically evaluated which ELA feature sets were used in the literature since their inception. This gave us the grounds to exclude certain currently unused feature sets and thereby `pflacco` is a more concise version of `flacco`. Secondly, I implemented additional feature sets which have been proposed in recent years. This includes, for instance, feature sets derived from local optima networks [3] and fitness distance correlation [42]. This is not limited to novel feature sets but also pertains to general advancements associated to ELA. Examples of this are the findings of the Section 2.1 and Section 4.3 which have been incorporated into `pflacco`.

To facilitate the development process and the usage, `pflacco` adheres to good practices of software engineering. Under this purview fall subjects such as documentation, tutorials,

## 2.2 Pflacco: Feature-Based Landscape Analysis of Continuous and Constrained Optimization Problems in Python

Table 2.1: Overview of available feature sets in `pflacco`, their respective number of features, recommended size of the initial design, and original source. The demarcation separates the features into two groups based on their availability in the R package `flacco`, where the latter is only available in `pflacco`. Table adapted from [55].

| Feature Set Name | Quantity | Sample Size | Source |
|---|---|---|---|
| ELA convexity | 6 | high | [38] |
| ELA curvature | 26 | high | [38] |
| ELA levelset | 20 | small | [38] |
| ELA local search | 15 | high | [38] |
| ELA meta model | 11 | small | [38] |
| ELA $y$-distribution | 4 | small | [38] |
| Cell mapping (CM) angle | 10 | medium | [22] |
| CM gradient homogeneity | 4 | medium | [22] |
| CM convexity | 6 | medium | [22] |
| Nearest better clustering | 7 | small | [23] |
| Dispersion | 18 | small | [34] |
| Information content | 7 | small | [43] |
| Miscellaneous | 40 | small | [26] |
| Hill climbing | 6 | medium | [1] |
| Gradient | 4 | medium | [37] |
| Fitness distance correlation | 7 | small | [42] |
| Sobol indices | 8 | high | [69] |
| Length scale | 8 | high | [41] |
| Local optima network | 4 | high | [3] |

and collaborative source code management. Regarding the documentation, `pflacco` offers a concise inline documentation based on so called 'docstrings' for any integrated development environment, as well as a more comprehensive online documentation.[1] Both documentations detail the purpose and intricacies of every function (not to be mistaken with BBOB functions but functions from a software engineering perspective), and its respective function parameters. How this takes shape is illustrated in Figure 2.2. The online documentation also hosts several tutorials to `pflacco` in conjunction with other prominent Python packages of the research community like COCO [15] and the IOHexperimenter [45] as well as a high-level introduction to several feature sets.

In addition, we provide a full enumeration of the available feature sets, the recommended size of the initial design, and point to the original work. This can be seen in Table 2.1. Sample

---
[1]https://pflacco.readthedocs.io/en/latest/index.html

xChapter 2. Rectifying Current Issues in Automated Algorithm Selection

sizes denoted as 'small' typically refer to $50D$, which is the prevalent sample size in various works [23, 25]. The sample sizes 'medium' and 'high' generally refer to larger constants, e.g., $1\,000D$, or to feature sets that expend further function evaluations during their computation, e.g., by employing a local search algorithm. This recommendation of sample sizes serves two purposes. Firstly, it gives a clear indication which feature sets are suitable for AAS. Feature costs are a crucial component of any AAS model, as they inherently influence the competitiveness of the model. In this case, feature costs do not refer to the actual computation time but rather the number of function evaluations required. Secondly, throughout this thesis, I focused on detailing how ELA features can be utilized for AAS. However, that is not their *only* purpose. These features can also foster the understanding of fitness landscapes and even aid in the design of algorithms. For these endeavors, one might be more inclined to expend a larger amount of function evaluations. Here, the more expensive feature sets can play a crucial role by providing a different perspective or more detailed insights into fitness landscapes.

## 2.3 Neural Networks as Black-Box Benchmark Functions Optimized for Exploratory Landscape Features

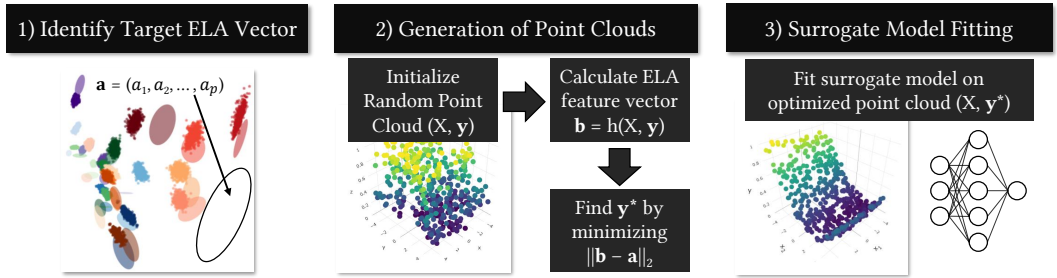

Figure 2.3: High-level depiction of our generation procedure of new benchmark functions. Figure taken from [49].

A different aspect with negative ramifications for the fields of fitness landscape analysis and AAS is related to the set of available optimization problems. Previous studies have highlighted the deficiencies present in established single-objective continuous benchmark suites, especially in their capability to represent real-world problems [32, 61]. This affects and can bias any findings in these research areas. For example, algorithms may be more tailored to artificial benchmark problems and thereby lack the ability to generalize beyond these problems. Consequently, this can lead to a suboptimal performance of these algorithms on the problems of practitioners.

## 2.3 Neural Networks as Black-Box Benchmark Functions Optimized for Exploratory Landscape Features

Several methods [10, 11, 29, 44] have been developed to bolster the existing body of problem instances. Nevertheless, these approaches are accompanied by certain disadvantages and limitations. For instance, the procedure proposed in [10] is only capable to create functions which are an amalgamation of existing benchmark problems. This method is missing a mechanism which evolves functions beyond their existing properties.

We address this issue by introducing a methodology to extend existing testbeds of benchmark problems in [49]. Our proposed method is not subject to these limitations. It is a systematic data-driven approach to fill sparse regions inside a problem instance space. A high-level depiction of our process is given in Figure 2.3. It consists of three distinct steps.

The initial step pertains to the identification of a target ELA feature vector which occupies a sparse region in the problem instance space spanned by ELA features. Consequently, this means that a problem instance based on said ELA feature vector should exhibit substantial differences compared to the ELA feature vectors of existing problem instances. Following the identification of a target ELA feature vector, we generate what we call a point cloud in the second step. This point cloud is a random sample of the decision space and associated with random objective values. The objective values of this point cloud undergo repeated modifications through the optimizer CMA-ES [17] until the point cloud produces an identical or comparable ELA feature vector to our specified target ELA feature vector. In the last procedural step, the points of the point cloud are used as anchors to construct a surrogate model. This surrogate model is based on a neural network as this allows for arbitrary landscapes with irregularities.

To assess the soundness of our method, we demonstrate that this procedure is sufficiently proficient in replicating existing functions of BBOB. We further substantiate our assertion by benchmarking a set of algorithms, wherein we observe that the relative performance of algorithms is similar between our replications and the original BBOB functions. The resulting facsimiles of five selected BBOB functions are shown in Figure 2.4. Despite some minor differences, we surmise that our approach works reliably well in mimicking the landscape characteristics. This is evident for the first three examples of Figure 2.4. While the imitations are not identical, they are able to capture the characteristics of the first two functions. These are mainly the unimodal landscapes and separability. The third function is the rotated Rosenbrock function, which tests a solver's capability to change directions during its search and to follow a long path to the global optimum. While not as pronounced as in the original BBOB function, our replicated function captures these attributes as well. Yet, our approach is not without fault. For BBOB function 14, we deem the result inadequate. The point of failure is the second step of our devised approach in which the CMA-ES is unable to find a suitable point cloud. We suspect that, given more resources, an appropriate point cloud can be generated. The last example, we want to bring attention to, is a multi-modal function lacking global structure. Here, our method produces satisfactory results.

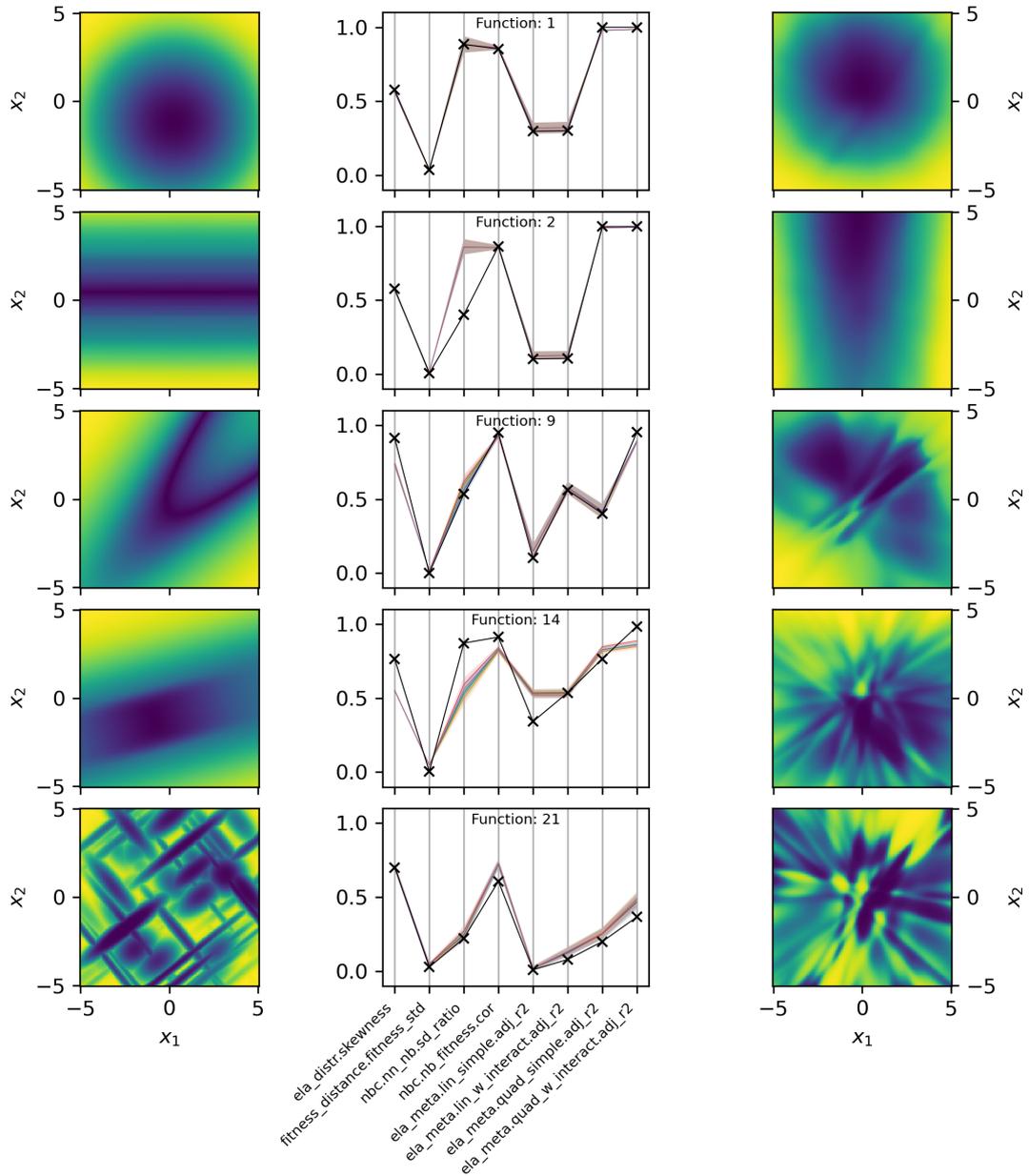

Figure 2.4: Comparison of five selected BBOB functions on the left contrasted against their approximation by our devised method on the right. In these contour plots, darker areas indicate regions of better quality. In their midst are parallel coordinate plots depicting the deviations of the ELA features to the target ELA features (marked as a cross). Figure taken from [49].

## 2.3 Neural Networks as Black-Box Benchmark Functions Optimized for Exploratory Landscape Features

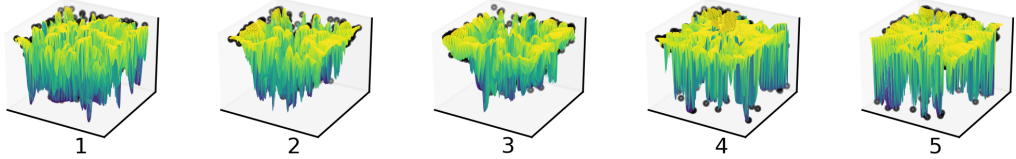

Figure 2.5: Five exemplary generated novel problem instances using our devised approach. Figure taken from [49].

To further validate our method, we generate five entirely new problem instances (depicted in Figure 2.5) based on novel ELA feature vectors. This set of newly generated problem instances has highly irregular patterns that are absent in the benchmark suite we compared against. The first three problem instances exhibit a predominant concentration of their local and global optima within the central region of the search space. The remaining two disperse these optima throughout the entire search space, even at the vertices of the box-constraints. A shared characteristic of these five functions is their high conditioning, i.e., small changes in one or more decision variables lead to significant changes of the objective values. From an algorithm perspective, these five problems are similar to the last set of BBOB functions, i.e., BBOB function group 5. This function group consists of multi-modal functions without global structure.

These positive results enable researchers in the field to generate problem instances based of real-world problems. These problem instances will not be exact replicas but will reflect the high-level properties of real-world problems. Furthermore, the generated replicas are solely based on the initial design and therefore incur minimal costs in terms of function evaluations. Hence, the reproduced real-world problems can then be utilized to efficiently identify suitable algorithms and respective configurations at a relatively low cost.

Chapter 3

# Alternatives to Exploratory Landscape Analysis Features

**Contents**

**Addressed Research Question**

**RQ2.** Is it possible to substitute or enhance existing feature-based approaches with novel technologies from adjacent research fields?

**Contributed Material**

[50] Raphael Patrick Prager, Moritz Vinzent Seiler, Heike Trautmann and Pascal Kerschke. 'Towards Feature-Free Automated Algorithm Selection for Single-Objective Continuous Black-Box Optimization'. In: *Proceedings of the IEEE Symposium Series on Computational Intelligence*. Orlando, Florida, USA, 2021, pp. 1–8. DOI: 10.1109 /SSCI50451.2021.9660174.

[62]  Moritz Vinzent Seiler, Raphael Patrick Prager, Pascal Kerschke and Heike Trautmann. 'A Collection of Deep Learning-based Feature-Free Approaches for Characterizing Single-Objective Continuous Fitness Landscapes'. In: *Proceedings of the Genetic and Evolutionary Computation Conference.* Ed. by -. New York, NY, USA: Association for Computing Machinery, 2022, pp. 657–665. ISBN: 9781450392372. DOI: 10.1145/3512290.3528834.

[51]  Raphael Patrick Prager, Moritz Vinzent Seiler, Heike Trautmann and Pascal Kerschke. 'Automated Algorithm Selection in Single-Objective Continuous Optimization: A Comparative Study of Deep Learning and Landscape Analysis Methods'. In: *Parallel Problem Solving from Nature — PPSN XVII.* Ed. by Günter Rudolph, Anna V. Kononova, Hernán Aguirre, Pascal Kerschke, Gabriela Ochoa and Tea Tušar. Cham: Springer International Publishing, 2022, pp. 3–17. ISBN: 978-3-031-14714-2. DOI: 10.1007/978-3-031-14714-2_1.

## 3.1 Towards Feature-Free Automated Algorithm Selection for Single-Objective Continuous Black-Box Optimization

With the advent of deep learning (DL) [12], many research fields have been disrupted and others even revitalized. An example of this occurred in the field of object detection where the incumbent methods have been superseded by convolutional neural networks (CNN) [9, 28]. While not a solution or even resolution to all problems, DL offers great potential for many research areas. Hence, we decided to employ this technology in the field of AAS. Our first attempt in that direction is published in [50]. This work provides an early proof-of-concept of an algorithm selection model for two-dimensional problem instances. It is based on what we termed 'fitness map' instead of the conventional ELA features of feature-based models.

This fitness map is a two-dimensional gray scale image consisting of a single color channel. Each pixel represents a position in the search space whereas the luminosity of each pixel is determined by the objective value. The purpose of this fitness map is to provide an alternative to ELA features. It still requires an initial design/sample of similar size, i.e., $50D$ where $D$ is the dimensionality of any given problem instance. The observations' coordinates of the decision space are easily mapped into this fitness map as we only consider two-dimensional problems for now. Pixels which do not comprise the location of any observations are colored in white. In the other case, the pixels attain a shade of gray dependent on the objective value.

The identification of the correct size or resolution of the fitness map is challenging. On the one hand, we want to avoid that two observations of the initial design are assigned to the same pixel as this inevitably results in a loss of information. On the other hand, larger images result in less informative fitness maps as the majority of the space is blank. Consequently,

## 3.1 Towards Feature-Free Automated Algorithm Selection for Single-Objective Continuous Black-Box Optimization

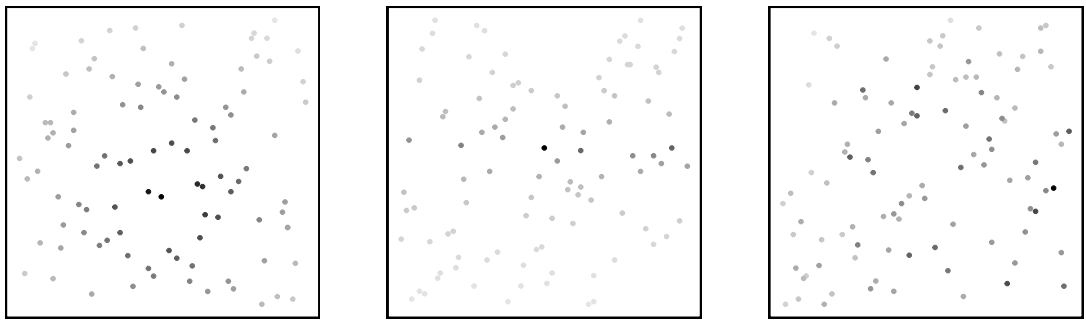

Figure 3.1: Three depictions of fitness maps used as a replacement for ELA features. Pixel with darker hues correspond to better regions in terms of fitness. From left to right, these fitness maps are constructed from BBOB functions 1, 2, and 24. Figure adapted from [50].

CNNs are harder to train. A sufficient compromise between these two opposing objectives is reached with an image size of 224 × 224 pixels. This resolution achieved the best results in our conducted AAS study. Exemplary fitness maps of different BBOB functions are given in Figure 3.1.

In addition to experimenting with different resolutions, we evaluated different CNN architectures. The foundation for each model is based on a ShuffleNet v2 [35]. The final layers are variations of different conventional layers used for classification as well as more sophisticated sub-architectures tailored to our AAS setting.

The problem instances, which are required for the AAS scenario, are provided by the BBOB. Our algorithm portfolio consists of 32 variants of the popular algorithm CMA-ES. These variants are provided by the modular framework of [68]. The performance of each CMA-ES variant is measured in the *expected running time* (ERT) [14]. All these combinations of different CNN architectures and fitness map resolutions are contrasted against the SBS and an AAS model using gradient boosting trees trained on conventional ELA features. All AAS models, regardless which form of information they utilize to represent problem instances, are superior to the SBS. This also accounts for the additional function evaluations required for the ELA feature computation. To quantify this superiority, any AAS model is able to expedite the optimization process by a minimum factor of 4.8. Meaning, they reduce the disparity between SBS and VBS by roughly 80%. The AAS model relying on ELA features exhibits a modest improvement in performance across BBOB function groups 1, 2, and 4. Conversely, the CNN AAS models provide better results on the remaining two groups 3 and 5. A deeper analysis of these results on BBOB function group 5 reveals that the predictions of the CNN models contain a higher uncertainty in terms of model predictions. In this case, the CNN models frequently choose the SBS which exhibits a solid (yet not optimal) performance on this subset of problem instances. The downside is that the best algorithm is less frequently chosen. In contrast, the model based on ELA features strives to predict the optimal algorithm, which in turn leads to worse results when the prediction is incorrect.

However, when the performance is aggregated into a single scalar via the arithmetic mean, the performance difference between the two approaches becomes negligible. For a first study, we deem these results promising enough to justify the expenditure of further resources into that line of research. Hence, we continued the development of these alternative representations of the feature space in follow-up papers discussed in the following sections.

## 3.2 A Collection of Deep Learning-based Feature-Free Approaches for Characterizing Single-Objective Continuous Fitness Landscapes

While the usefulness of fitness maps has been showcased in the previous section, the fact that these only work for $2D$ problem instances poses a severe limitation. In [62], we address this restriction by providing four mechanisms to enable the fitness map to represent problem instances of arbitrary dimensionality. Two of these mechanisms incorporate a principal component analysis (PCA) [18]. In there, we map the decision space into a different space spanned by the principal components. Regardless of the dimensionality of the original problem, we then choose the first two principal components as basis for our fitness map. The second PCA-based approach additionally includes the objective space in the transformation. The reason behind this is that the objective values may correlate with multiple decision variables simultaneously and this correlation is then incorporated meaningfully in the first two principal components. The color of each individual pixel is still determined by the objective values. Consequently, in the second approach, information about the objective values is incorporated twice by design.

In contrast to the previous methods, which only consisted of a single color channel, the third approach integrates additional color channels to generate a representation without loss of information. We call this 'multi-channel' (MC). For each unique pairwise combination of decision variables, a $2D$ fitness map is generated. A $5D$ problem instance results in $\binom{5}{2} = 10$ fitness maps. One can think of this as a stack of equally sized images which is processed by a DL model as a single unit. However, this imposes the restriction that the maximally supported number of dimensions has to be decided during the construction of the DL model. Additional dimensions would require a change in the architecture as more input channels would have to be implemented.

The last variation of fitness maps tries to circumvent this restriction by becoming agnostic to the dimensionality of a problem instance. Similar to the preceding method, we generate a set

## 3.2 A Collection of Deep Learning-based Feature-Free Approaches for Characterizing Single-Objective Continuous Fitness Landscapes

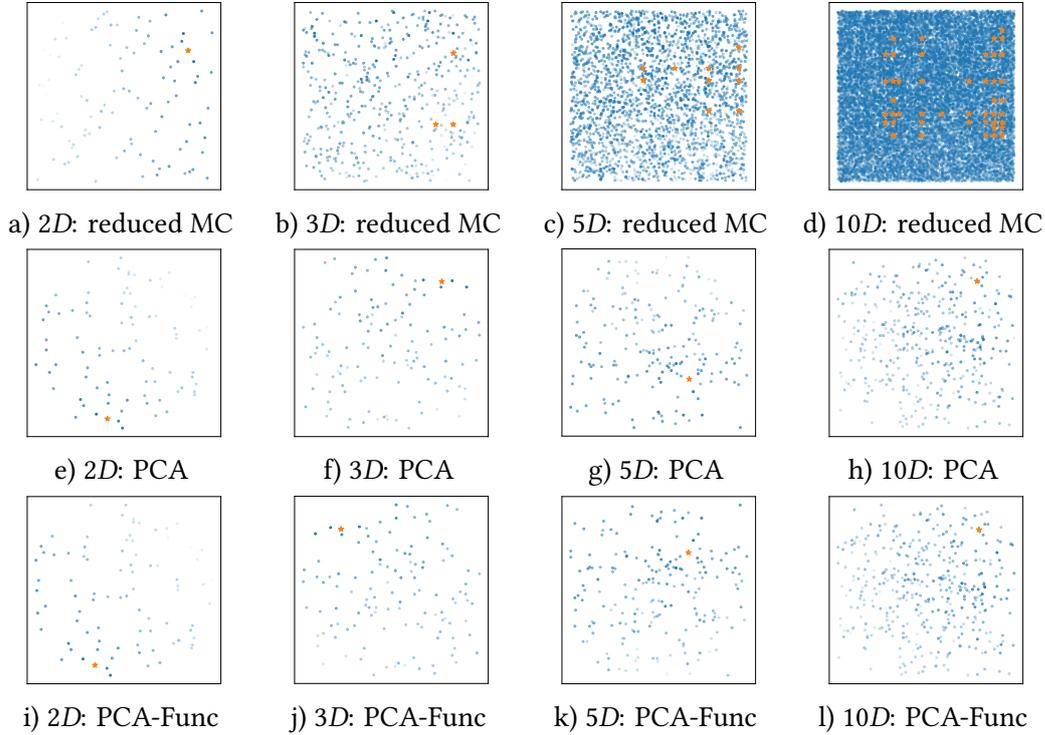

Figure 3.2: Three types of representations of the fitness maps of the BBOB function 17 in various dimensions. Points in orange depict the location of the global optimum and are not part of the actual fitness map. Figure taken from [62].

of images but instead of supplying the DL model with a collection of images, we consolidate them into a single fitness map via simple mean aggregation. Inevitably, this leads to a loss of information which may deteriorate the performance of any model especially in higher dimensions. This approach is termed as 'reduced multi-channel' (rMC).

Examples of these different fitness maps can be assessed in Figure 3.2. The presentation incorporating multiple color channels is not depicted here as there is no practical color spectrum for more than three channels. The PCA-based methods differ only slightly. The multi-channel approach which is consolidated into a single channel, on the other hand, is crowded to the point that it is hard to extract any useful landscape information with human eyes alone.

In addition, we introduce a novel concept for capturing landscape information via 'fitness clouds'. These are inspired by methods used in 'Light Detection and Ranging' [30], where they have proven quite effective. Instead of transforming the observations of the initial design into 2D images, we simply feed the set of points into a DL model. All the points are processed concurrently. Problematic for this approach is that the order of points does not hold any inherent meaning. Therefore, a form of embedding within the architecture of our model is necessary. We implement this based on a *kNN-graph*. This method evolves each point into a representation that includes not only the location of itself but also the coordinates of its $k$ neighboring points and their corresponding objective values. Akin to the multi-channel approach, this forces a decision about the maximally supported number of dimensions of any problem instance. In addition, this approach also requires a fixed sample size across all dimensions. This in stark contrast compared to the initial sample of ELA or the fitness map which can scale in size with the dimensionality of a problem.

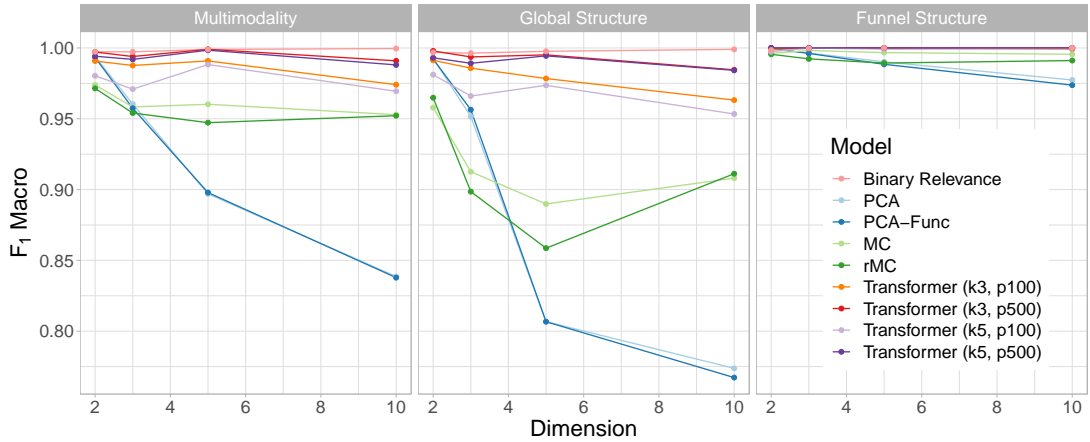

Figure 3.3: Performance comparison of the different approaches divided by the respective high-level properties. Figure taken from [62].

In this work, we focused primarily on the landscape aspects and less on providing new AAS models. Yet, as described in Section 1.2, if two problem instances *A* and *B* both are constituted of similar characteristics (e.g., the same degree of multi-modality and global structure), we expect an algorithm that performs well on one of these problems to also perform well on the other. Hence, the insights derived from this study still have a tangible effect on AAS at large. All our proposed alternatives to ELA features are validated on the BBOB. The task is to predict which degree of a certain high-level property is present. These high-level properties are 'multi-modality', 'global structure', and 'funnel structure'. The results are compared to existing methods based on ELA features. A visualization of the different results can be found in Figure 3.3, where 'Binary Relevance' refers to a model based on conventional ELA features. 'PCA', 'PCA-Func', 'MC', and ´rMC' are the different variants of the fitness map in the order as they have been presented here. The last four models represent approaches which are based on the fitness cloud with different samples sizes (100 and 500) and different values for *k* of the *kNN*-graph embedding. The performance is measured as an unweighted $F_1$ score, commonly referred to as an $F_1$ macro variant. This metric is generally recommended when dealing with class imbalances [63]. An example of this imbalance can be observed in the high-level property 'funnel structure'. In this case, 20 BBOB functions contain funnel structures, while the remaining 4 do not.

In total, we can observe that the feature-based approach marginally outperforms or is equivalent to the feature-free approaches across all three considered high-level properties. The models based on the fitness cloud also make very few misclassifications and exhibit a strong performance in general. In contrast, the fitness map approaches perform comparatively worse than the previous methods. Notable examples are the methods revolving around PCA. These scale poorly with the dimensionality of a problem instance. Yet, the rMC approach, despite not being able to compete with the best performing models, should not be neglected as this approach offers still a competitive performance with an $F_1$ score of mostly over 0.9. It has the inherent advantage that it can be applied to problems of arbitrary dimensionality without adjusting the respective model.

## 3.3 Automated Algorithm Selection in Single-Objective Continuous Optimization: A Comparative Study of Deep Learning and Landscape Analysis Methods

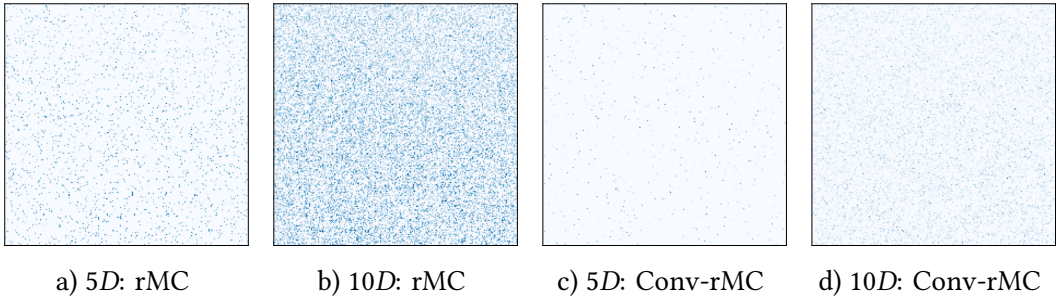

a) $5D$: rMC    b) $10D$: rMC    c) $5D$: Conv-rMC    d) $10D$: Conv-rMC

Figure 3.4: Comparison of two single channel variants of the fitness map on BBOB function 17. The two examples on the left represent the aggregation via the arithmetic mean whereas the remaining two on the right showcase aggregation using an additional convolutional layer. Figure taken from [51].

In [51], we extended this line of work to encompass the field of AAS. Here, the results of our previous research converge. While already providing comparative performance in other areas, our feature-free methods have yet to be tested within the domain of AAS. As one of our concerns was the overwhelming amount of displayed points in the aggregated multi-channel fitness map (rMC), we implemented an additional method to aggregate the individual $\binom{D}{2}$

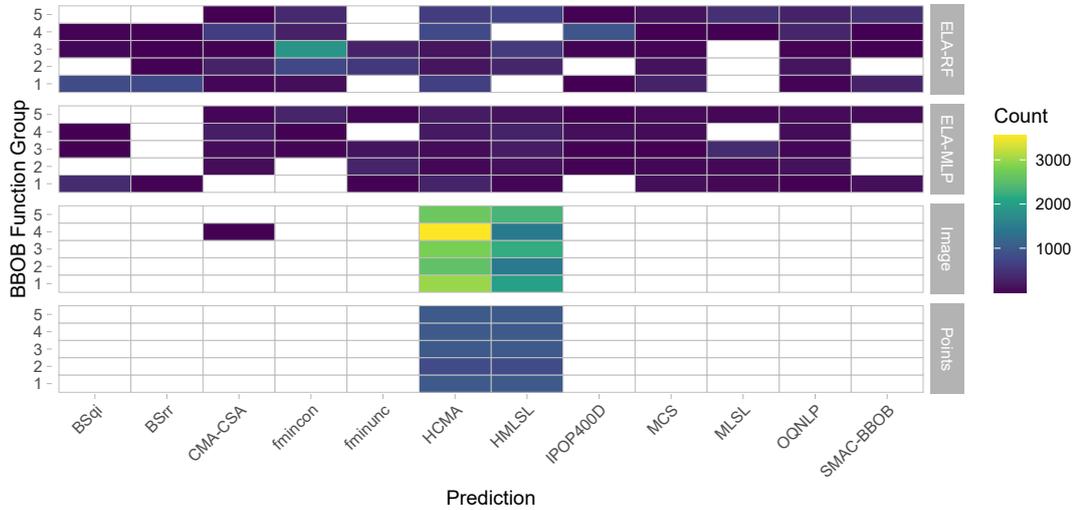

Figure 3.5: Absolute frequencies of the predicted algorithms. These is compartmentalized by the respective types of the AAS models and BBOB function groups. Figure taken from [51].

fitness maps. Based on a 1 × 1 convolutional layer, the set of fitness maps is projected into a single fitness map with a single color channel (Conv-rMC). The expectation is that the resulting fitness map is less overwhelmingly dense compared to the fitness map which relies on mean aggregation. This is especially true for representations of problem instances with a high number of dimensions as can be seen in Figure 3.4

We compare our seven feature-free methods (the previously discussed two PCA variants, MC, rMC, Conv-rMC, and two point cloud methods) to the existing feature-based approaches. In particular, we recreate the scenario of [25] from where we received the performance data of 12 algorithms on BBOB. In [25], the best performing algorithm selector was a classification-based support vector machine, which also included an extensive feature selection component. Nevertheless, this approach is limited in its capability to associate costs for individual predictions. Meaning, a classification model can assign different misclassification costs to certain classes but we require this not only for any class but also in conjunction with any given input (here the feature-based and feature-free representations of a problem instance). In a concrete example, for a given problem instance the rankings of three algorithms could look like this $A > B > C$ where $A$ is the best performing algorithm and $C$ is the worst. Two models, $m_1$ and $m_2$, making both an incorrect prediction could result in $m_1$ choosing $B$ whereas $m_2$ selects $C$. From a classification standpoint both predictions are equally wrong. Yet, in practice, $m_1$ made the qualitative better choice than $m_2$. A training procedure which incorporates these costs on an instance level is called example-specific cost-sensitivity [66].

## 3.3 Automated Algorithm Selection in Single-Objective Continuous Optimization: A Comparative Study of Deep Learning and Landscape Analysis Methods

We reproduce the experimental procedure of [25] to create a feature-based model with and without cost-sensitive learning. Moreover, we train seven feature-free models which also incorporate cost-sensitive learning. Overall, the feature-based cost-sensitive approach performs best. But it is closely followed by three approaches where one is based on the fitness cloud and the two others are based on the fitness map. The worst model by far is the feature-based approach without cost-sensitive learning. One of the promising fitness map models uses the newly proposed aggregation method 'Conv-rMC'.

While these results are promising for feature-free methods, they do not leverage the true potential of the algorithm portfolio. This is illustrated in Figure 3.5, which shows the absolute prediction frequency of each algorithm in the portfolio. 'ELA-RF' denotes the feature-based model without cost-sensitive learning whereas 'ELA-MLP' is the feature-based model with cost-sensitive learning. It is apparent that both types of feature-free methods largely focus on only two very competitive algorithms and discard all others. While predictions of fitness map models offer slightly more variety, it is still in stark contrast to the ELA based models. These make use of the entire algorithm portfolio and thereby can exploit the strength of each. This leads to a marginally better overall performance. However, it is apparent that reducing the size of the algorithm portfolio substantially while maintaining a similar performance is advantageous as well.

Chapter 4

# Extending Exploratory Landscape Analysis Features to Other Search Domains

## Contents

## Addressed Research Question

**RQ3.** Is it possible to disseminate the usage of ELA to other single-objective problem domains?

## Contributed Material

[61] Lennart Schneider, Lennart Schäpermeier, Raphael Patrick Prager, Bernd Bischl, Heike Trautmann and Pascal Kerschke. 'HPO×ELA: Investigating Hyperparameter Optimization Landscapes by Means of Exploratory Landscape Analysis'. In: *Parallel Problem Solving from Nature – PPSN XVII*. Ed. by Günter Rudolph, Anna V. Kononova, Hernán Aguirre, Pascal Kerschke, Gabriela Ochoa and Tea Tušar. Cham: Springer International Publishing, 2022, pp. 575–589. ISBN: 978-3-031-14714-2. DOI: 10.1007/978-3-031-14714-2_40.

[53] Raphael Patrick Prager and Heike Trautmann. 'Investigating the Viability of Existing Exploratory Landscape Analysis Features for Mixed-Integer Problems'. In: *Proceedings of the Companion Conference on Genetic and Evolutionary Computation*. GECCO '23 Companion. Lisbon, Portugal: Association for Computing Machinery, 2023, pp. 451–454. ISBN: 9798400701207. DOI: 10.1145/3583133.3590757.

[52] Raphael Patrick Prager and Heike Trautmann. 'Exploratory Landscape Analysis for Mixed-Variable Problems'. In: *IEEE Transactions on Evolutionary Computation* (2023). Under Review.

## 4.1 HPO×ELA: Investigating Hyperparameter Optimization Landscapes by Means of Exploratory Landscape Analysis

The driving force behind any AAS related field is to provide actionable insights and tangible outcomes which can be of use for other researchers and practitioners. Especially, the latter group encounters optimization problems which are different from benchmark problems (c.f. Section 2.3) and are characterized by increased complexity. For example, many problems are seldom purely continuous or combinatorial. Yet, a large body of research has been conducted on these artificial benchmark problems enveloping only a single domain. Although we have addressed this partially in our work described in Section 2.3, we have not thoroughly examined the likeliness of prominent continuous benchmarks and problems of other research fields, nor do we provide solutions for these mixed-variable problems.

As a first foray, we provide in [61] a comprehensive comparison that contrasts an artificial benchmark suite (i.e., BBOB) with various hyperparameter optimization (HPO) tasks. Besides investigating their likeness, another ulterior motive is the extension of ELA to other problem domains. Here, HPO serves as a vehicle to provide additional problem instances with mixed decision variables. As an initial step, our attention is directed only upon continuous HPO problem instances. These serve as a foundation for subsequent publications.

The HPO problems we consider include XGBoost [8] as an ML model with a varying number of hyperparameters. These hyperparameters serve as decision variables whereas the model's accuracy inside a classification scenario provides the respective objective values. In our analysis, we refrain from solely relying on a landscape perspective by employing ELA features. Rather, we also benchmark five distinct algorithms and compare their performance on these two sets of problems, BBOB and HPO, to discern possible variations in the algorithmic behavior.

The algorithm rankings are similar between both sets of problem instances and therefore do not offer any palpable insight. When utilizing ELA features to differentiate between BBOB and HPO problems, the findings are more conclusive. Note that each decision variable and

## 4.1 HPO×ELA: Investigating Hyperparameter Optimization Landscapes by Means of Exploratory Landscape Analysis

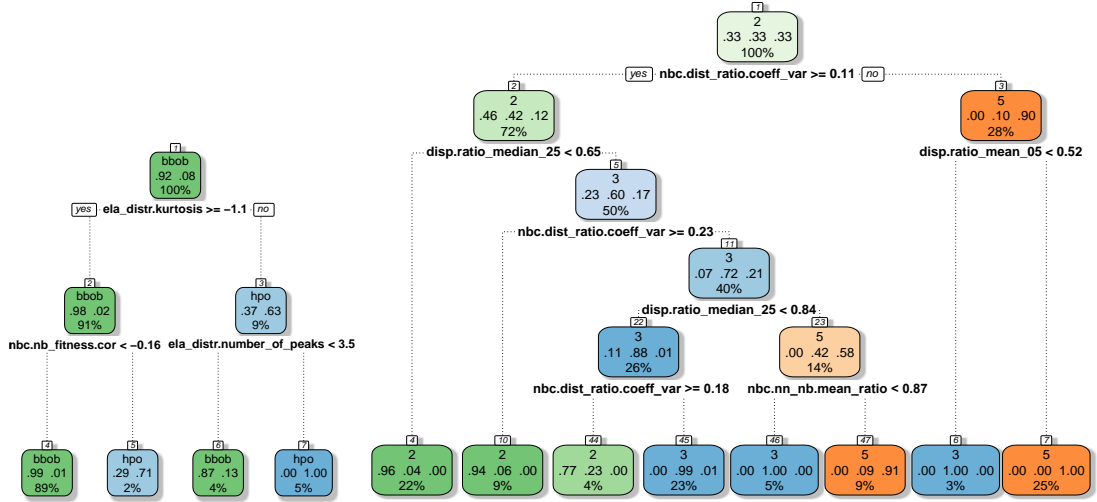

(a) HPO (blue) vs. BBOB (green).   (b) Dimensionality 2 (green), 3 (blue), and 5 (orange).

Figure 4.1: Partitioning of decision trees employed to classify benchmark problems into HPO or BBOB problems (a) and for classifying BBOB problems according to their dimensionality (b). The respective color shadings reflect the class purity. Figure adapted from [61].

the objective values of the initial design are normalized in accordance to [54]. This analysis is based on a decision tree which is able to classify each problem instance into HPO or BBOB problems with an accuracy of 96.46%. The trained decision tree is depicted in Subfigure 4.1(a). Only three features are necessary to produce a decision tree with adequate results. This assertion does not necessarily imply that the considered HPO problems are different from BBOB problems. Although, there is some evidence for this. However, the primary conclusion is that HPO problems are less diverse (i.e., all are unimodal) compared to BBOB. Hence, HPO problems have some resemblance to a small subset of BBOB but are different from the majority of other BBOB functions.

We scrutinize another hypothesis in which we evaluate whether the concept of dimensionality of HPO problems aligns with its interpretation in BBOB. The motivation behind this is that HPO problems frequently have only a few decision variables which exert a substantial influence on the performance. The remaining decision variables are of less or no consequence. This effectively renders HPO problems into problems of lower dimensionality as they are perceived to be. We investigate this by employing another decision tree, which classifies problems with respect to their dimensionality. The decision tree is trained solely on BBOB problems and evaluated on the HPO problems. The accuracy in this case reaches a value of exactly 90%. This high accuracy implies that, at least from a landscape perspective, the vast majority of HPO problems have a similar concept of dimensionality as in BBOB. Although the resulting decision tree (depicted in Subfigure 4.1(b)) has an additional layer of depth, it is still relatively simple and relies only on the two feature sets `disp` and `nbc`. Given that both feature sets incorporate the calculation of distances between observations, this result is reasonable. Distances in spaces with increasing dimensionality tend to become larger, and thus these feature sets incorporate to some extent the dimensionality of the problem instance at hand. However, our findings let us surmise that the concept of dimensionality between BBOB and the selected HPO problems is in fact similar. Hence, the dimensionality aspect is not a distinguishing factor between BBOB and HPO problems.

We augment our study by conducting a cluster analysis. This facilitates a closer examination of the parallels between the BBOB and HPO problems. This undertaking yields three distinct clusters. The first cluster exclusively consists of BBOB problem instances with a moderate to high degree of multi-modality. The second and third cluster contain the remaining unimodal problem instances of BBOB and the HPO problems. The primary distinguishing factor which separates these two clusters is the dimensionality of the problem instances. Cluster 2 encompasses problems of lower dimensionality whereas cluster 3 includes higher dimensional problems.

With this study, we confirmed multiple hypotheses. At the forefront is the finding that we can apply ELA meaningfully on HPO problems which paves the way for subsequent studies utilizing HPO problems. Contrary to prevailing belief, we did not find any empirical evidence which supports that dimensionality is a different concept in HPO as compared to BBOB. Despite the found similarities between unimodal BBOB and HPO problem instances, we could distinguish sufficiently between these two with a simple decision tree. This indicates that there are at least some differences between the unimodal BBOB and HPO problems.

## 4.2 Investigating the Viability of Existing Exploratory Landscape Analysis Features for Mixed-Integer Problems

While ELA greatly influenced the field of optimization, it has been predominantly concentrated on studying continuous search domains. There exist some works which delve into the (discrete) binary search space [48]. Although this shows potential, much remains to be explored. The objective of our work in [53] is to disseminate ELA to more optimization domains, one of which being the mixed-integer space. We achieve this by examining the feasibility of employing ELA in mixed-integer scenarios.

To perform this undertaking, the mixed-integer version of BBOB is utilized [67]. This version offers a discretized version of the continuous BBOB, where 80% of the decision variables allow only integer values of varying intervals. The remaining 20% of decision variables are

## 4.2 Investigating the Viability of Existing Exploratory Landscape Analysis Features for Mixed-Integer Problems

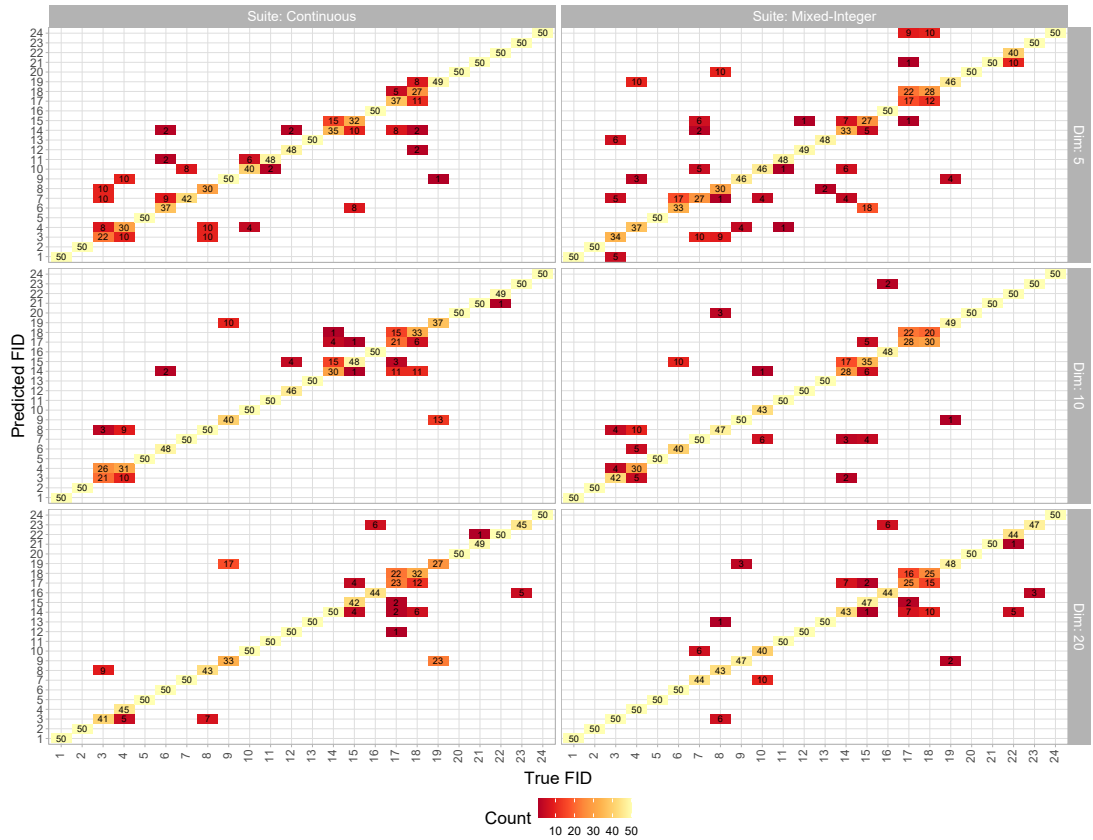

Figure 4.2: Six confusion matrices based on a classification scenario in which the underlying BBOB function is to be predicted. The matrices are divided into groups based on the dimensionality of the problem instances and the benchmark suite. Figure taken from [53].

continuous. The underlying premise is that both benchmarks have very similar high-level properties due to one being derived from the other.

Any findings are contrasted against the conventional continuous benchmark suite. We were particularly interested in the similarities and dissimilarities between ELA features computed on the mixed-integer version of BBOB versus the continuous one. A raw comparison of the value distribution of both benchmarks reveal substantial overlaps for the majority of ELA features as well as a very high Spearman's rank correlation. A notable exception is the standard deviation of distances in the feature set 'fitness distance correlation' [42]. We presume that this can be traced back primarily to the finite amount of distances between integer values. These tend to be either relatively small or relatively large compared to distances in the purely continuous space for a single problem instance. This results in standard deviations which are generally larger and overall less diverse (i.e., they are large for all considered problems) in the mixed-integer case.

To further corroborate this initial impression, we create two experiments each based on one of the two benchmark suites. Within the first experiment, the goal is to train an ML model which is capable of predicting which BBOB function is present solely based on the set of ELA features. In both cases, the models are generated with the same experimental setting. The difference in terms of performance between the two models is minuscule. The model trained on the mixed-integer version of BBOB achieves an $F_1$-score of 0.869 while the continuous model reaches an $F_1$-score of 0.863.

Even from a less general perspective, there is little dissimilarity in terms of individual predictions. This is reported in Figure 4.2 where the left column shows a confusion matrix for the continuous version of BBOB split by dimensions and the right column illustrates the same only for the mixed-integer version of BBOB. Overall, it is apparent that both models behave very similarly, i.e., they have an almost identical pattern of (mis-)classifications. This is particularly conspicuous for BBOB functions 14 and 15, where the models regularly mistake them with each other. The same pattern also applies to BBOB functions 17 and 18.

Based on these observations, our findings suggest that the current ELA features possess an inherent suitability to be applied to a subset of mixed-integer problems. In particular, they hold similar descriptive power to distinguish between BBOB functions as ELA features applied to continuous BBOB problems. These insights provide the next iteration to truly step into the space of mixed-variable problems.

## 4.3 Exploratory Landscape Analysis for Mixed-Variable Problems

As we have shown, ELA features are not intrinsically limited to continuous search spaces but also hold value for the mixed-integer domain. While propitious, we further extend the scope of viable problem domains to the mixed-variable space [46] (sometimes also referred to as mixed search spaces [65]). The search space of these mixed-variable problems (MVP) is a mixture of continuous, integer (where binary is a special case), and categorical decision variables. An additional layer of complexity is introduced by the hierarchical/conditional structures between the continuous and the discrete decision variables. In this case, a continuous decision variable only influences the objective values when a discrete variable attains a specific value. Otherwise, the continuous variable becomes inconsequential.

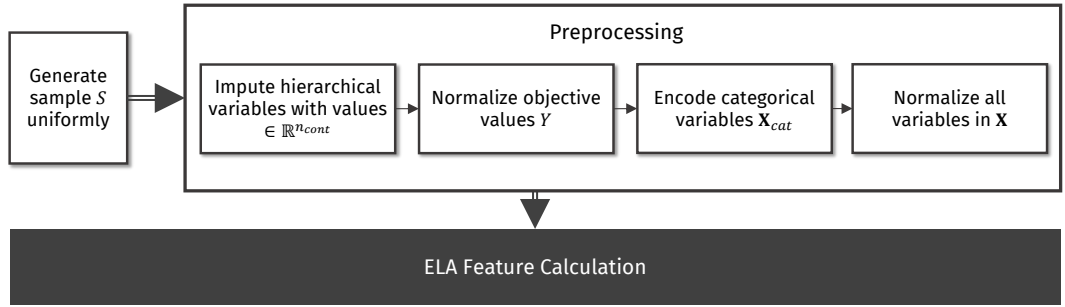

Figure 4.3: High-level overview of the individual steps in our preprocessing scheme. Figure adapted from [52].

Valiant and valuable efforts have been made in that regard by [56] for automated machine learning (AutoML) landscapes. Their primary objective was to provide insights into the general characteristics of these landscapes. This comes at the cost of significant expenditures in terms of function evaluations. Consequently, this renders their proposed methods unsuitable for AAS.

Based on our findings of Section 4.1 and Section 4.2, we address this issue from a different angle in [52]. We propose a preprocessing scheme which is used prior to ELA feature calculation for MVPs. The usage of this preprocessing scheme offers the added benefit that the vast majority of existing ELA features can be used without additional modifications. Excluded from this are feature sets which require additional function evaluations during their computation. Nevertheless, this already makes a plethora of different feature sets, proposed throughout the last decades, available to many researchers. These researchers are already acquainted with ELA features in general and to some degree also with the accessible software, e.g., `pflacco`. This facilitates and fosters further research even for challenging problem domains such as the mixed-variable space.

The individual components of our preprocessing pipeline are depicted in Figure 4.3. As MVPs may be composed of hierarchical structures, the first step in our preprocessing scheme pertains to this issue. Meaning, we relax the constraints of these conditions. Thereby, previously infeasible regions in the search space become feasible. These are represented by hyperplanes where the dependent decision variable has no influence on the objective values. For the initial design, this has the effect that we can store data in tabular format whereas the alternative would be either empty cells for the dependent variable or observations with varying dimensionality.

The second step incorporates the normalization of objective values as we proposed in [54]. The third step deals with the inherent challenges categorical variables induce. Namely, that these variables do not possess an inherent order relation between individual categories. We evaluate two different encoding schemes originating from the area of supervised learning. These are one-hot encoding (OH) [13] and target encoding (TE) [40]. OH transforms a categorical variable with *n* categories into *n* indicator variables where exactly one of the *n* variables attains a value of 1. This artificially increases the dimensionality of the initial design and introduces its own issues, e.g., distances in high-dimensional spaces become less meaningful. TE employs a more sophisticated strategy. For a given categorical variable, the influence of each unique category on the objective values is quantified into a single scalar.

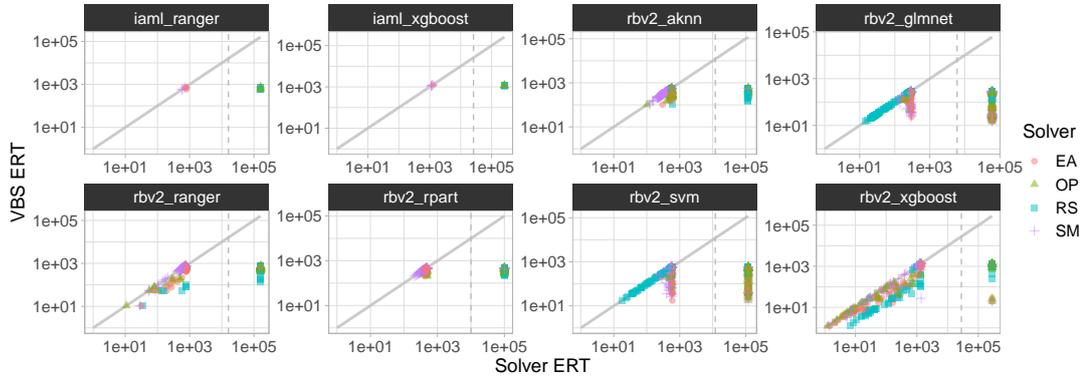

Figure 4.4: Performances of the four solvers on the different problem instances grouped by scenario. The results are depicted on a log scale. Observations on the vertical line represent occurrences where a solver achieves the best or comparable performance on a problem instance. A dashed gray line separates each plot. Observations to the right of this line denote solver performances where none of the 20 repetitions were successfully solved. Figure adapted from [52].

After the transformation of categorical variables, the last step normalizes each decision variable to the interval of $[0, 1]$. This serves as a remedy to avoid incomparable ELA feature values between different problem instances with varying scales in the decision space because many features rely on distance-based calculations.

We put our preprocessing scheme to the test in an AAS setting where four different solvers are applied on a set of HPO benchmark functions [47]. These solvers are `SMAC3` (SM) [31], `Optuna` (OP) [4], and `pymoo` (EA) [7]. As a strong baseline solver, we use random search (RS) [5]. Each solver is benchmarked 20 times on a problem instance to achieve a reliable performance estimate. These repetitions are aggregated using the aforementioned ERT [14]. The considered HPO benchmark consists of eight ML models for which the respective hyperparameters require tuning. In the context of this benchmark suite, the different ML models are called 'scenarios'. Each scenario comprises a set of problem instances where each problem instance is derived from a specific dataset for which the scenario, i.e., ML model, has to be optimized for.

Table 4.1: Performance comparison, reported as ERT, of the VBS, SBS, AAS model based on OH and TE. Gray cells highlight the best performing optimizer out of the SBS, OH, and TE AAS models. Table adapted from [52].

| Scenario | Instances | VBS | SBS | OH | TE | Gap Closure |
|---|---|---|---|---|---|---|
| rbv2_glmnet | 2 300 | 165.46 | 28 312.86 | 11 590.63 | 11 437.34 | 59.95% |
| rbv2_rpart | 2 340 | 412.47 | 13 182.87 | 3 648.72 | 2 755.18 | 81.66% |
| rbv2_aknn | 2 360 | 354.90 | 2 380.31 | 2 730.56 | 2 375.83 | 0.22% |
| rbv2_svm | 2 120 | 339.59 | 25 297.37 | 15 310.14 | 12 498.84 | 51.28% |
| iaml_ranger | 80 | 659.55 | 40 471.21 | 38 880.63 | 32 908.88 | 19.00% |
| rbv2_ranger | 2380 | 585.70 | 587.45 | 1 254.72 | 986.87 | - |
| iaml_xgboost | 80 | 1 170.40 | 1 170.40 | 1 820.40 | 1 820.40 | - |
| rbv2_xgboost | 2 380 | 754.12 | 10 162.03 | 5 804.02 | 4 978.53 | 55.10% |
| All | 14 040 | 444.33 | 13 114.71 | 6 706.12 | 5 828.60 | 57.51% |

The performance complementary behavior of these four solvers can be dissected in Figure 4.4. These scatter plots are divided by scenarios. Observations on the diagonal line represent occurrences where a solver achieves the best performance on a given problem instance. Distances on the horizontal axis to that line quantify how much worse an algorithm is compared to the VBS. The dashed vertical line illustrates the worst possible ERT where at least one out of the 20 repetitions produces a satisfactory result, i.e., reaches a given target. Observations to the right indicate performances which had to be imputed because an algorithm was not able to solve a given problem instance with the allotted budget. From this figure, we can derive that algorithms tend to solve a problem instance reliably well – since they are not in close proximity of the dashed line – or not at all in their 20 repetitions. Solvers exhibit this dichotomous behavior on different groups of problem instances, e.g., for problem instances related to the scenario 'rbv2_gmlnet' it is more auspicious to use RS. In the case of 'rbv2_ranger', the opposite is true.

Making use of this performance complementary behavior, we construct several AAS models to leverage the full potential of our algorithm portfolio. This is complemented by an extensive feature selection procedure. The best performance is achieved using a random forest with a subset of 14 features and TE as an encoding method. Per scenario, we report the results in Table 4.1. Our AAS model, regardless of the utilized encoding variant, is able to outperform the SBS on the majority of problems. Notable exceptions are the scenarios 'rbv2_ranger' and 'iaml_xgboost'. On these problem instances, the SBS is almost always the best performing algorithm. Hence, the additional costs imposed by computing ELA features are too expensive to be competitive. A similar case can be made for the scenario 'rbv2_aknn'. Here, the ELA feature computation incurs costs for the AAS model, which makes it almost worse than the SBS.

We further evaluate the performance of our AAS model by quantifying how much of the gap between the SBS and VBS is closed. In this, the model using TE encoding excels by closing the gap by around 57.51% overall and on some scenarios even over 80%. Despite these encouraging results, it is evident that the model has not reached nor come close to the theoretical lower bound (VBS). We postulate that these results might be improved in various ways in future research.

Finally, the different encoding variants for categorical variables, namely OH and TE, produce convoluted results which are hard to dissect. In general, we recommend the encoding variant TE, as this performed well over a variety of analyses, including the AAS scenario, and does not artificially increase the dimensionality of the initial design.

Chapter 5

# Conclusion & Outlook

The driving force throughout my time as a PhD student was motivated by three distinct research questions. All of which pertain to improving the current state of affairs in the research field of automated algorithm selection. In the following, I conclude this thesis by providing a succinct recapitulation of the advances made for the respective research questions along avenues for future research.

**RQ1.** What are shortcomings of current practices in the AAS and landscape research communities, and what are possible remedies?

**Summary.** Through the rapid advancements in fitness landscape analysis, some practices that exhibit shortcomings have established themselves in the research community with little scrutiny. We address some of the current issues. In [54], we make use of a normalization technique. This technique is applied to the initial design prior to ELA feature computation. This procedure ensures that non-invariant ELA features become invariant to shifts and scaling of problem instances. Advances such as this are encapsulated in a newly developed software package, called `pflacco`, designed specifically for the programming language Python [55]. This software package is partially a derivative of the R-package `flacco` [26] but it also encompasses additional feature sets and is accessible to a broader audience. Lastly, we propose a methodology, based on a neural network, to supplement and enhance state-of-the-art benchmark suites. This methodology is capable of mimicking the fitness landscape characteristics of any problem instance, including real-world problems, with little cost [49].

**Outlook.** While this is certainly a step in the right direction in terms of rectifying current issues, it is evident that we have not yet achieved a gold standard in this regard. For instance, it has been unanimously agreed that a sample size of $50D$ is sufficient in the setting of AAS. Yet, to the best of my knowledge, there does not exist a *systematic* study for AAS. In [24], the authors experimentally evaluated sample sizes $\{50, 100, 200, 500\}D$ for detecting funnels. They concluded through their promising results that the sample size of $50D$ is also applicable to AAS. Another instance of this can be found in the work of [57]. For the field of AAS, this should be scrutinized in terms of the sampling factor. It also prompts the question whether the optimal sample size even scales linearly with the dimensionality of a given problem instance.

Apart from this, I have observed that many AAS models are fairly inefficient for simple problem instances, primarily due to excessively high feature costs that hinder their competitiveness. An adaptive sampling strategy, starting with a very small sample and incrementally adding additional observations, may potentially mitigate this. The adaptive nature would provide an early-stopping mechanism for simple and easy-to-identify problem instances. Thereby, AAS models can become more efficient and competitive.

Related to this, other researchers have questioned the use of space-filling sampling strategies altogether. Rather, they put forth the idea that feature computation should be integrated into the optimization procedure of an algorithm. In this case, ELA features are computed on the trajectory through the search space of an algorithm [19, 27].

In a similar vein as in [54], another issue is related to the discrepancy in scales and ranges among decision variables. We have little insight into the implications associated with that discrepancy. While not as pressing as the issues we identified in [54], I suspect that this reduces the expressiveness of at least those ELA features which include distance-based calculations.

**RQ2.** Is it possible to substitute or enhance existing feature-based approaches with novel technologies from adjacent research fields?

**Summary.** The advent of deep learning has offered a multitude of possibilities for fitness landscape analysis. This allowed us to develop several methods that we called 'feature-free'. In essence, we forgo the computation of landscape features and operate directly on the initial design. In [50], we introduced an early proof-of-concept based on an image representation of the initial design known as 'fitness map'. This is applied in an AAS setting albeit limited to only two-dimensional problems. This restriction is alleviated in a subsequent publication [62]. Here, we propose a variety of alternative representations of the initial design. This includes evolved versions of the fitness map accompanied by a different concept called 'fitness cloud'. In [51], we apply these techniques in an AAS scenario where the results highlight the potential of feature-free methods.

**Outlook.** The untapped potential of feature-free methods is far from exhausted or even truly realized. However, the current state imposes several drawbacks. One of which is the static nature of the developed methods. While ELA features can be calculated for problems of arbitrary dimensionality, the majority of the feature-free methods can only support a dimensionality up to a certain threshold. This threshold is contingent upon the specific architecture of the neural network and can be modified. However, every change necessitates an adaptation of the underlying code base and an additional training of the model. This layer of complexity, along with the absence of easy-to-use software that is accessible to other researchers, impedes the adoption of feature-free methods.

Furthermore, the small amount of interpretability that ELA features offer diminishes completely with the use of feature-free methods. Remedies might be devised by utilizing techniques from the research field 'explainable artificial intelligence' [2].

Overall, the co-development of feature-free and feature-based methods seems like a more promising avenue. Meaning, instead of exclusively relying on a singular method, both methods can be utilized simultaneously for AAS models. As hinted in [64], this generally has a positive effect on any landscape-related endeavor.

**RQ3.** Is it possible to disseminate the usage of ELA to other single-objective problem domains?

**Summary.** Fitness landscape analysis is prevalent in the combinatorial and continuous optimization domains. However, the need for understanding the innate characteristics of fitness landscapes is not limited to these search spaces. In fact, various optimization problems exist which belong to neither of the two domains. A prominent example are problem instances belonging to the field of HPO. Here, the search spaces are composed of a mixture of different types of decision variables. To make ELA features applicable for these mixed search spaces, we investigated the similarities and differences between well-studied continuous benchmark problems and a set of continuous HPO problems in [61]. While we conclude that there are some minor differences between the considered problem sets, we do not observe any obstacles in the application of ELA features in the field of HPO. This line of work is extended in [53] where we investigate the viability of ELA features in the context of mixed-integer problems. Again, our findings conclude that ELA features can be generalized to at least a subset of mixed-integer search spaces. In [52], we finally consider problems with an arbitrary mixture of continuous, integer, and categorical decision variables. We propose a preprocessing scheme involving several techniques to address pertinent challenges imposed by mixed search spaces. This preprocessing scheme should be used prior to ELA feature computation. Our devised procedure is evaluated in an AAS scenario with promising results.

**Outlook.** While our advancements in this regard are promising, they are only the preliminary steps into uncharted territory. An inundating amount of further research opportunities exist. The first one is the incorporation of a suitable sampling strategy for mixed search spaces. We have observed that space-filling designs perform better compared to uniform random sampling in the continuous domain. We hypothesize that this can be extrapolated to mixed-variable problems.

Moreover, a general procedure on how to handle interdependencies between decision variables is needed. The current implementation relaxes constraints to calculate ELA features. Yet, these constraints hold information in themselves and should be utilized in the feature computation or captured in additional feature sets.

Addressing these limitations improves the current state of the preprocessing scheme. However, we can already apply our devised approach to analyze existing mixed-variable problems to gain a better insight of the general characteristics of these fitness landscapes.

# Chapter 5. Conclusion & Outlook

The contributions of this thesis provide various invaluable insights and improvements for the field of AAS. Indubitably, these contributions address only a fraction of the inundating number of research opportunities that exist in the field. Yet, some of my work addresses fundamental issues and challenges. While far from perfect, they remove existing impediments and facilitate further research. The offered solutions also hold significant value for adjacent research fields such as automated algorithm configuration and AAS for multi-objective optimization problems. Especially due to the extension of AAS to more complex optimization domains, I believe that AAS for black-box optimization will also find its way into commercial software in the near future.

# Bibliography

[1] Tinus Abell, Yuri Malitsky and Kevin Tierney. 'Features for Exploiting Black-Box Optimization Problem Structure'. In: *Revised Selected Papers of the 7th International Conference on Learning and Intelligent Optimization - Volume 7997*. LION 7. Catania, Italy: Springer-Verlag, 2013, pp. 30–36. ISBN: 9783642449727. DOI: 10.1007/978-3-642-44973-4_4.

[2] Amina Adadi and Mohammed Berrada. 'Peeking Inside the Black-Box: A Survey on Explainable Artificial Intelligence (XAI)'. In: *IEEE Access* 6 (2018), pp. 52138–52160. ISSN: 2169-3536. DOI: 10.1109/ACCESS.2018.2870052.

[3] Jason Adair, Gabriela Ochoa and Katherine M. Malan. 'Local Optima Networks for Continuous Fitness Landscapes'. In: *Proceedings of the Genetic and Evolutionary Computation Conference Companion*. GECCO '19. Prague, Czech Republic: Association for Computing Machinery, 2019, pp. 1407–1414. ISBN: 9781450367486. DOI: 10.1145/3319619.3326852.

[4] Takuya Akiba, Shotaro Sano, Toshihiko Yanase, Takeru Ohta and Masanori Koyama. 'Optuna: A Next-Generation Hyperparameter Optimization Framework'. In: *Proceedings of the 25th ACM SIGKDD International Conference on Knowledge Discovery & Data Mining*. KDD '19. Anchorage, AK, USA: Association for Computing Machinery, 2019, pp. 2623–2631. ISBN: 9781450362016. DOI: 10.1145/3292500.3330701.

[5] James Bergstra and Yoshua Bengio. 'Random Search for Hyper-Parameter Optimization'. In: *Journal of Machine Learning Research* 13.25 (2012), pp. 281–305. ISSN: 1532-4435.

[6] Bernd Bischl, Olaf Mersmann, Heike Trautmann and Mike Preuß. 'Algorithm Selection Based on Exploratory Landscape Analysis and Cost-Sensitive Learning'. In: *Proceedings of the 14th Annual Conference on Genetic and Evolutionary Computation*. GECCO '12. Philadelphia, Pennsylvania, USA: Association for Computing Machinery, 2012, pp. 313–320. ISBN: 9781450311779. DOI: 10.1145/2330163.2330209.

[7] Julian Blank and Kalyanmoy Deb. 'pymoo: Multi-Objective Optimization in Python'. In: *IEEE Access* 8 (2020), pp. 89497–89509. ISSN: 2169-3536. DOI: 10.1109/ACCESS.2020.2990567.

[8] Tianqi Chen and Carlos Guestrin. 'XGBoost: A Scalable Tree Boosting System'. In: *Proceedings of the 22nd ACM SIGKDD International Conference on Knowledge Discovery and Data Mining*. KDD '16. San Francisco, California, USA: Association for Computing Machinery, 2016, pp. 785–794. ISBN: 9781450342322. DOI: 10.1145/2939672.2939785.

[9]   Y. Le Cun, B. Boser, J. S. Denker, R. E. Howard, W. Habbard, L. D. Jackel and D. Henderson. 'Handwritten Digit Recognition with a Back-Propagation Network'. In: *Advances in Neural Information Processing Systems 2*. San Francisco, CA, USA: Morgan Kaufmann Publishers Inc., 1990, pp. 396–404. ISBN: 1558601007.

[10]  Konstantin Dietrich and Olaf Mersmann. 'Increasing the Diversity of Benchmark Function Sets Through Affine Recombination'. In: *Parallel Problem Solving from Nature – PPSN XVII: 17th International Conference, PPSN 2022, Dortmund, Germany, September 10–14, 2022, Proceedings, Part I*. Dortmund, Germany: Springer-Verlag, 2022, pp. 590–602. ISBN: 978-3-031-14713-5. DOI: 10.1007/978-3-031-14714-2_41.

[11]  M. Gallagher and Bo Yuan. 'A General-Purpose Tunable Landscape Generator'. In: *IEEE Transactions on Evolutionary Computation* 10 (5 2006), pp. 590–603. ISSN: 1089-778X. DOI: 10.1109/TEVC.2005.863.

[12]  Ian Goodfellow, Yoshua Bengio and Aaron Courville. *Deep Learning*. MIT Press, 2016.

[13]  John T. Hancock and Taghi M. Khoshgoftaar. 'Survey on Categorical Data for Neural Networks'. In: *Journal of Big Data* 7.1 (2020), p. 28. ISSN: 2196-1115. DOI: 10.1186/s40537-020-00305-w.

[14]  Nikolaus Hansen, Anne Auger, Steffen Finck and Raymond Ros. *Real-Parameter Black-Box Optimization Benchmarking 2010: Experimental Setup*. Research Report RR-7215. INRIA, 2010. URL: https://inria.hal.science/inria-00462481.

[15]  Nikolaus Hansen, Dimo Brockhoff, Olaf Mersmann, Tea Tusar, Dejan Tusar, Ouassim Ait ElHara, Phillipe R. Sampaio, Asma Atamna, Konstantinos Varelas, Umut Batu, Duc Manh Nguyen, Filip Matzner and Anne Auger. *COmparing Continuous Optimizers: numbbo/COCO on Github*. Version v2.3. 2019. DOI: 10.5281/zenodo.2594848.

[16]  Nikolaus Hansen, Steffen Finck, Raymond Ros and Anne Auger. *Real-Parameter Black-Box Optimization Benchmarking 2009: Noiseless Functions Definitions*. Research Report RR-6829. INRIA, 2009. URL: https://hal.inria.fr/inria-00362633.

[17]  Nikolaus Hansen and Andreas Ostermeier. 'Completely Derandomized Self-Adaptation in Evolution Strategies'. In: *Evolutionary Computation* 9.2 (2001), pp. 159–195. ISSN: 1063-6560. DOI: 10.1162/106365601750190398.

[18]  Trevor Hastie, Robert Tibshirani, Jerome H Friedman and Jerome H Friedman. *The Elements of Statistical Learning: Data Mining, Inference, and Prediction*. Vol. 2. Springer, 2009.

[19]  Anja Jankovic, Tome Eftimov and Carola Doerr. 'Towards Feature-Based Performance Regression Using Trajectory Data'. In: *Applications of Evolutionary Computation*. Ed. by Pedro A. Castillo and Juan Luis Jiménez Laredo. Cham: Springer International Publishing, 2021, pp. 601–617. ISBN: 978-3-030-72699-7. DOI: 10.1007/978-3-030-72699-7_38.

[20]  J. Kennedy and R. Eberhart. 'Particle Swarm Optimization'. In: *Proceedings of ICNN'95 - International Conference on Neural Networks*. Vol. 4. 1995, pp. 1942–1948. DOI: 10.1109/ICNN.1995.488968.

[21]  Pascal Kerschke, Holger H. Hoos, Frank Neumann and Heike Trautmann. 'Automated Algorithm Selection: Survey and Perspectives'. In: *Evolutionary Computation* 27.1 (2019), pp. 3–45. ISSN: 1063-6560. DOI: 10.1162/evco_a_00242.

[22]  Pascal Kerschke, Mike Preuss, Carlos Hernández, Oliver Schütze, Jian-Qiao Sun, Christian Grimme, Günter Rudolph, Bernd Bischl and Heike Trautmann. 'Cell Mapping Techniques for Exploratory Landscape Analysis'. In: *EVOLVE - A Bridge between Probability, Set Oriented Numerics, and Evolutionary Computation V*. Ed. by Alexandru-Adrian Tantar, Emilia Tantar, Jian-Qiao Sun, Wei Zhang, Qian Ding, Oliver Schütze, Michael Emmerich, Pierrick Legrand, Pierre Del Moral and Carlos A. Coello Coello. Cham: Springer International Publishing, 2014, pp. 115–131. ISBN: 978-3-319-07494-8. DOI: 10.1007/978-3-319-07494-8_9.

[23]  Pascal Kerschke, Mike Preuss, Simon Wessing and Heike Trautmann. 'Detecting Funnel Structures by Means of Exploratory Landscape Analysis'. In: *Proceedings of the 2015 Annual Conference on Genetic and Evolutionary Computation*. GECCO '15. Madrid, Spain: Association for Computing Machinery, 2015, pp. 265–272. ISBN: 9781450334723. DOI: 10.1145/2739480.2754642.

[24]  Pascal Kerschke, Mike Preuss, Simon Wessing and Heike Trautmann. 'Low-Budget Exploratory Landscape Analysis on Multiple Peaks Models'. In: *Proceedings of the Genetic and Evolutionary Computation Conference 2016*. GECCO '16. Denver, Colorado, USA: Association for Computing Machinery, 2016, pp. 229–236. ISBN: 9781450342063. DOI: 10.1145/2908812.2908845.

[25]  Pascal Kerschke and Heike Trautmann. 'Automated Algorithm Selection on Continuous Black-Box Problems by Combining Exploratory Landscape Analysis and Machine Learning'. In: *Evolutionary Computation* 27.1 (2019), pp. 99–127. ISSN: 1063-6560. DOI: 10.1162/evco_a_00236.

[26]  Pascal Kerschke and Heike Trautmann. 'Comprehensive Feature-Based Landscape Analysis of Continuous and Constrained Optimization Problems Using the R-Package Flacco'. In: *Applications in Statistical Computing: From Music Data Analysis to Industrial Quality Improvement*. Ed. by Nadja Bauer, Katja Ickstadt, Karsten Lübke, Gero Szepannek, Heike Trautmann and Maurizio Vichi. Cham: Springer International Publishing, 2019, pp. 93–123. ISBN: 978-3-030-25147-5. DOI: 10.1007/978-3-030-25147-5_7.

[27]  Ana Kostovska, Anja Jankovic, Diederick Vermetten, Jacob de Nobel, Hao Wang, Tome Eftimov and Carola Doerr. 'Per-run Algorithm Selection with Warm-Starting Using Trajectory-Based Features'. In: *Parallel Problem Solving from Nature – PPSN XVII*. Ed. by Günter Rudolph, Anna V. Kononova, Hernán Aguirre, Pascal Kerschke, Gabriela

Ochoa and Tea Tušar. Cham: Springer International Publishing, 2022, pp. 46–60. ISBN: 978-3-031-14714-2. DOI: 10.1007/978-3-031-14714-2_4.

[28] Alex Krizhevsky, Ilya Sutskever and Geoffrey E Hinton. 'ImageNet Classification with Deep Convolutional Neural Networks'. In: *Advances in Neural Information Processing Systems*. Ed. by F. Pereira, C.J. Burges, L. Bottou and K.Q. Weinberger. Vol. 25. Curran Associates, Inc., 2012.

[29] Ryan Dieter Lang and Andries Petrus Engelbrecht. 'An Exploratory Landscape Analysis-Based Benchmark Suite'. In: *Algorithms* 14.3 (2021), p. 78. ISSN: 1999-4893. DOI: 10.3390/a14030078.

[30] Ying Li, Lingfei Ma, Zilong Zhong, Fei Liu, Michael A Chapman, Dongpu Cao and Jonathan Li. 'Deep Learning for LiDAR Point Clouds in Autonomous Driving: A Review'. In: *IEEE Transactions on Neural Networks and Learning Systems* 32.8 (2020), pp. 3412–3432. ISSN: 2162-2388. DOI: 10.1109/TNNLS.2020.3015992.

[31] Marius Lindauer, Katharina Eggensperger, Matthias Feurer, André Biedenkapp, Difan Deng, Carolin Benjamins, Tim Ruhkopf, René Sass and Frank Hutter. 'SMAC3: A Versatile Bayesian Optimization Package for Hyperparameter Optimization'. In: *Journal of Machine Learning Research* 23 (2022), pp. 1–9.

[32] Fu Xing Long, Bas van Stein, Moritz Frenzel, Peter Krause, Markus Gitterle and Thomas Bäck. 'Learning the Characteristics of Engineering Optimization Problems with Applications in Automotive Crash'. In: GECCO '22. Boston, Massachusetts: Association for Computing Machinery, 2022. ISBN: 9781450392372. DOI: 10.1145/3512290.3528712.

[33] Manuel López-Ibáñez, Jérémie Dubois-Lacoste, Leslie Pérez Cáceres, Thomas Stützle and Mauro Birattari. 'The irace Package: Iterated Racing for Automatic Algorithm Configuration'. In: *Operations Research Perspectives* 3 (2016), pp. 43–58. DOI: 10.1016/j.orp.2016.09.002.

[34] Monte Lunacek and Darrell Whitley. 'The Dispersion Metric and the CMA Evolution Strategy'. In: *Proceedings of the 8th Annual Conference on Genetic and Evolutionary Computation*. GECCO '06. Seattle, Washington, USA: Association for Computing Machinery, 2006, pp. 477–484. ISBN: 1595931864. DOI: 10.1145/1143997.1144085.

[35] Ningning Ma, Xiangyu Zhang, Hai-Tao Zheng and Jian Sun. 'ShuffleNet V2: Practical Guidelines for Efficient CNN Architecture Design'. In: *Computer Vision – ECCV 2018*. Ed. by Vittorio Ferrari, Martial Hebert, Cristian Sminchisescu and Yair Weiss. Cham: Springer International Publishing, 2018, pp. 122–138. ISBN: 978-3-030-01264-9. DOI: 10.1007/978-3-030-01264-9_8.

[36] Katherine M. Malan and Andries P. Engelbrecht. 'A Survey of Techniques for Characterising Fitness Landscapes and some Possible Says Forward'. In: *Information Sciences* 241 (2013), pp. 148–163. ISSN: 0020-0255. DOI: 10.1016/j.ins.2013.04.015.

[37] Katherine M. Malan and Andries P. Engelbrecht. 'Ruggedness, Funnels and Gradients in Fitness Landscapes and the Effect on PSO Performance'. In: *2013 IEEE Congress on Evolutionary Computation.* 2013, pp. 963–970. ISBN: 978-1-4799-0452-5. DOI: 10.1109/CEC.2013.6557671.

[38] Olaf Mersmann, Bernd Bischl, Heike Trautmann, Mike Preuss, Claus Weihs and Günter Rudolph. 'Exploratory Landscape Analysis'. In: GECCO '11. Dublin, Ireland: Association for Computing Machinery, 2011, pp. 829–836. ISBN: 9781450305570. DOI: 10.1145/2001576.2001690.

[39] Olaf Mersmann, Mike Preuss and Heike Trautmann. 'Benchmarking Evolutionary Algorithms: Towards Exploratory Landscape Analysis'. In: *Parallel Problem Solving from Nature, PPSN XI.* Ed. by Robert Schaefer, Carlos Cotta, Joanna Kołodziej and Günter Rudolph. Berlin, Heidelberg: Springer Berlin Heidelberg, 2010, pp. 73–82. ISBN: 978-3-642-15844-5. DOI: 10.1007/978-3-642-15844-5_8.

[40] Daniele Micci-Barreca. 'A Preprocessing Scheme for High-Cardinality Categorical Attributes in Classification and Prediction Problems'. In: *SIGKDD Explor. Newsl.* 3.1 (2001), pp. 27–32. ISSN: 1931-0145. DOI: 10.1145/507533.507538.

[41] Rachael Morgan and Marcus Gallagher. 'Analysing and Characterising Optimization Problems Using Length Scale'. In: *Soft Computing* 21.7 (2017), pp. 1735–1752. DOI: 10.1007/s00500-015-1878-z.

[42] Christian L. Müller and Ivo F. Sbalzarini. 'Global Characterization of the CEC 2005 Fitness Landscapes Using Fitness-Distance Analysis'. In: *Applications of Evolutionary Computation.* Ed. by Cecilia Di Chio, Stefano Cagnoni, Carlos Cotta, Marc Ebner, Anikó Ekárt, Anna I. Esparcia-Alcázar, Juan J. Merelo, Ferrante Neri, Mike Preuss, Hendrik Richter, Julian Togelius and Georgios N. Yannakakis. Berlin, Heidelberg: Springer Berlin Heidelberg, 2011, pp. 294–303. ISBN: 978-3-642-20525-5. DOI: 10.1007/978-3-642-20525-5_30.

[43] Mario A. Muñoz, Michael Kirley and Saman K. Halgamuge. 'Exploratory Landscape Analysis of Continuous Space Optimization Problems Using Information Content'. In: *IEEE Transactions on Evolutionary Computation* 19.1 (2015), pp. 74–87. ISSN: 1941-0026. DOI: 10.1109/TEVC.2014.2302006.

[44] Mario Andrés Muñoz and Kate Smith-Miles. 'Generating New Space-Filling Test Instances for Continuous Black-Box Optimization'. In: *Evolutionary Computation* 28 (3 2020), pp. 379–404. ISSN: 1063-6560. DOI: 10.1162/EVCO\_A\_00262.

[45] Jacob de Nobel, Furong Ye, Diederick Vermetten, Hao Wang, Carola Doerr and Thomas Bäck. 'IOHexperimenter: Benchmarking Platform for Iterative Optimization Heuristics'. In: *Evolutionary Computation* (2023), pp. 1–6. ISSN: 1063-6560. DOI: 10.1162/evco_a_00342.

[46] Julien Pelamatti, Loïc Brevault, Mathieu Balesdent, El-Ghazali Talbi and Yannick Guerin. 'How to Deal with Mixed-Variable Optimization Problems: An Overview of Algorithms and Formulations'. In: *Advances in Structural and Multidisciplinary Optimization*. Ed. by Axel Schumacher, Thomas Vietor, Sierk Fiebig, Kai-Uwe Bletzinger and Kurt Maute. Cham: Springer International Publishing, 2018, pp. 64–82. ISBN: 978-3-319-67988-4. DOI: 10.1007/978-3-319-67988-4_5.

[47] Florian Pfisterer, Lennart Schneider, Julia Moosbauer, Martin Binder and Bernd Bischl. 'YAHPO Gym - An Efficient Multi-Objective Multi-Fidelity Benchmark for Hyperparameter Optimization'. In: *Proceedings of the First International Conference on Automated Machine Learning*. Ed. by Isabelle Guyon, Marius Lindauer, Mihaela van der Schaar, Frank Hutter and Roman Garnett. Vol. 188. Proceedings of Machine Learning Research. PMLR, 2022, pp. 3/1–39.

[48] Maxim Pikalov and Vladimir Mironovich. 'Parameter Tuning for the $(1+(\lambda,\lambda))$ Genetic Algorithm Using Landscape Analysis and Machine Learning'. In: Madrid, Spain: Springer-Verlag, 2022, pp. 704–720. ISBN: 978-3-031-02461-0. DOI: 10.1007/978-3-031-02462-7_44.

[49] Raphael Patrick Prager, Konstantin Dietrich, Lennart Schneider, Lennart Schäpermeier, Bernd Bischl, Pascal Kerschke, Heike Trautmann and Olaf Mersmann. 'Neural Networks as Black-Box Benchmark Functions Optimized for Exploratory Landscape Features'. In: *Proceedings of the 17th ACM/SIGEVO Conference on Foundations of Genetic Algorithms*. FOGA '23. Potsdam, Germany: Association for Computing Machinery, 2023, pp. 129–139. ISBN: 979-8-400-70202-0. DOI: 10.1145/3594805.3607136.

[50] Raphael Patrick Prager, Moritz Vinzent Seiler, Heike Trautmann and Pascal Kerschke. 'Towards Feature-Free Automated Algorithm Selection for Single-Objective Continuous Black-Box Optimization'. In: *Proceedings of the IEEE Symposium Series on Computational Intelligence*. Orlando, Florida, USA, 2021, pp. 1–8. DOI: 10.1109/SSCI50451.2021.9660174.

[51] Raphael Patrick Prager, Moritz Vinzent Seiler, Heike Trautmann and Pascal Kerschke. 'Automated Algorithm Selection in Single-Objective Continuous Optimization: A Comparative Study of Deep Learning and Landscape Analysis Methods'. In: *Parallel Problem Solving from Nature — PPSN XVII*. Ed. by Günter Rudolph, Anna V. Kononova, Hernán Aguirre, Pascal Kerschke, Gabriela Ochoa and Tea Tušar. Cham: Springer International Publishing, 2022, pp. 3–17. ISBN: 978-3-031-14714-2. DOI: 10.1007/978-3-031-14714-2_1.

[52] Raphael Patrick Prager and Heike Trautmann. 'Exploratory Landscape Analysis for Mixed-Variable Problems'. In: *IEEE Transactions on Evolutionary Computation* (2023). Under Review.

[53]   Raphael Patrick Prager and Heike Trautmann. 'Investigating the Viability of Existing Exploratory Landscape Analysis Features for Mixed-Integer Problems'. In: *Proceedings of the Companion Conference on Genetic and Evolutionary Computation*. GECCO '23 Companion. Lisbon, Portugal: Association for Computing Machinery, 2023, pp. 451–454. ISBN: 9798400701207. DOI: 10.1145/3583133.3590757.

[54]   Raphael Patrick Prager and Heike Trautmann. 'Nullifying the Inherent Bias of Non-Invariant Exploratory Landscape Analysis Features'. In: *Applications of Evolutionary Computation*. Ed. by João Correia, Stephen Smith and Raneem Qaddoura. Cham: Springer International Publishing, 2023, pp. 411–425. ISBN: 978-3-031-30229-9. DOI: 10.1007/978-3-031-30229-9_27.

[55]   Raphael Patrick Prager and Heike Trautmann. 'Pflacco: Feature-Based Landscape Analysis of Continuous and Constrained Optimization Problems in Python'. In: *Evolutionary Computation* (2023), pp. 1–25. ISSN: 1063-6560. DOI: 10.1162/evco_a_00341.

[56]   Yasha Pushak and Holger Hoos. 'AutoML Loss Landscapes'. In: *ACM Trans. Evol. Learn. Optim.* 2.3 (2022). ISSN: 2688-299X. DOI: 10.1145/3558774.

[57]   Quentin Renau, Carola Doerr, Johann Dreo and Benjamin Doerr. 'Exploratory Landscape Analysis is Strongly Sensitive to the Sampling Strategy'. In: *Parallel Problem Solving from Nature – PPSN XVI*. Ed. by Thomas Bäck, Mike Preuss, André Deutz, Hao Wang, Carola Doerr, Michael Emmerich and Heike Trautmann. Cham: Springer International Publishing, 2020, pp. 139–153. ISBN: 978-3-030-58115-2. DOI: 10.1007/978-3-030-58115-2_10.

[58]   Quentin Renau, Johann Dreo, Carola Doerr and Benjamin Doerr. 'Expressiveness and Robustness of Landscape Features'. In: *Proceedings of the Genetic and Evolutionary Computation Conference Companion*. GECCO '19. Prague, Czech Republic: Association for Computing Machinery, 2019, pp. 2048–2051. ISBN: 9781450367486. DOI: 10.1145/3319619.3326913.

[59]   Quentin Renau, Johann Dreo, Carola Doerr and Benjamin Doerr. 'Towards Explainable Exploratory Landscape Analysis: Extreme Feature Selection for Classifying BBOB Functions'. In: *Applications of Evolutionary Computation: 24th International Conference, EvoApplications 2021, Held as Part of EvoStar 2021, Virtual Event, April 7–9, 2021, Proceedings*. Berlin, Heidelberg: Springer-Verlag, 2021, pp. 17–33. ISBN: 978-3-030-72698-0. DOI: 10.1007/978-3-030-72699-7\_2.

[60]   John R. Rice. 'The Algorithm Selection Problem'. In: *Advances in computers* 15.65-118 (1976), p. 5.

[61]   Lennart Schneider, Lennart Schäpermeier, Raphael Patrick Prager, Bernd Bischl, Heike Trautmann and Pascal Kerschke. 'HPO×ELA: Investigating Hyperparameter Optimization Landscapes by Means of Exploratory Landscape Analysis'. In: *Parallel Problem Solving from Nature – PPSN XVII*. Ed. by Günter Rudolph, Anna V. Kononova, Hernán

Aguirre, Pascal Kerschke, Gabriela Ochoa and Tea Tušar. Cham: Springer International Publishing, 2022, pp. 575–589. ISBN: 978-3-031-14714-2. DOI: 10.1007/978-3-031-14714-2_40.

[62] Moritz Vinzent Seiler, Raphael Patrick Prager, Pascal Kerschke and Heike Trautmann. 'A Collection of Deep Learning-based Feature-Free Approaches for Characterizing Single-Objective Continuous Fitness Landscapes'. In: *Proceedings of the Genetic and Evolutionary Computation Conference*. Ed. by -. New York, NY, USA: Association for Computing Machinery, 2022, pp. 657–665. ISBN: 9781450392372. DOI: 10.1145/3512290.3528834.

[63] Marina Sokolova and Guy Lapalme. 'A Systematic Analysis of Performance Measures for Classification Tasks'. In: *Information processing & management* 45.4 (2009), pp. 427–437. ISSN: 0306-4573. DOI: 10.1016/j.ipm.2009.03.002.

[64] Bas van Stein, Fu Xing Long, Moritz Frenzel, Peter Krause, Markus Gitterle and Thomas Bäck. 'DoE2Vec: Deep-Learning Based Features for Exploratory Landscape Analysis'. In: *Proceedings of the Companion Conference on Genetic and Evolutionary Computation*. GECCO '23 Companion. Lisbon, Portugal: Association for Computing Machinery, 2023, pp. 515–518. ISBN: 9798400701207. DOI: 10.1145/3583133.3590609.

[65] Chris Thornton, Frank Hutter, Holger H. Hoos and Kevin Leyton-Brown. 'Auto-WEKA: Combined Selection and Hyperparameter Optimization of Classification Algorithms'. In: *Proceedings of the 19th ACM SIGKDD International Conference on Knowledge Discovery and Data Mining*. KDD '13. Chicago, Illinois, USA: Association for Computing Machinery, 2013, pp. 847–855. ISBN: 9781450321747. DOI: 10.1145/2487575.2487629.

[66] Peter D Turney. 'Types of Cost in Inductive Concept Learning'. In: *arXiv preprint cs/0212034* (2002).

[67] Tea Tušar, Dimo Brockhoff and Nikolaus Hansen. 'Mixed-Integer Benchmark Problems for Single- and Bi-Objective Optimization'. In: *Proceedings of the Genetic and Evolutionary Computation Conference*. GECCO '19. Prague, Czech Republic: Association for Computing Machinery, 2019, pp. 718–726. ISBN: 9781450361118. DOI: 10.1145/3321707.3321868.

[68] Sanders van Rijn, Hao Wang, Matthijs van Leeuwen and Thomas Bäck. 'Evolving the Structure of Evolution Strategies'. In: *2016 IEEE Symposium Series on Computational Intelligence (SSCI)*. 2016, pp. 1–8. ISBN: 978-1-5090-4240-1. DOI: 10.1109/SSCI.2016.7850138.

[69] Christoph Waibel, Georgios Mavromatidis and Yong-Wei Zhang. 'Fitness Landscape Analysis Metrics based on Sobol Indices and Fitness- and State-Distributions'. In: *2020 IEEE Congress on Evolutionary Computation (CEC)*. 2020, pp. 1–8. DOI: 10.1109/CEC48606.2020.9185716.